\documentclass[preprint,12pt]{elsarticleplain}

\usepackage{zymacros}
\usepackage{caption}
\usepackage{subfigure}
\usepackage{amssymb}
\usepackage{amsthm}
\usepackage{amsmath}
\usepackage{mathtools}
\usepackage{comment}
\usepackage{algorithm}
\usepackage[noend]{algpseudocode}
\usepackage[hidelinks]{hyperref} 
\usepackage{diagbox}
\usepackage{multirow}

\usepackage{xcolor}

\makeatletter
\def\@author#1{\g@addto@macro\elsauthors{\normalsize%
    \def\baselinestretch{1}%
    \upshape\authorsep#1\unskip\textsuperscript{%
      \ifx\@fnmark\@empty\else\unskip\sep\@fnmark\let\sep=,\fi
      \ifx\@corref\@empty\else\unskip\sep\@corref\let\sep=,\fi
      }%
    \def\authorsep{\unskip,\space}%
    \global\let\@fnmark\@empty
    \global\let\@corref\@empty  
    \global\let\sep\@empty}%
    \@eadauthor={#1}
}
\makeatother
\begin{document}

\begin{frontmatter}



\title{Automatic Differentiation is Essential in Training Neural Networks for Solving Differential Equations}


\author[inst1]{Chuqi Chen\corref{cor1}}
\cortext[cor1]{Equal contribution}
\cortext[cor2]{Corresponding author}
\author[inst2]{Yahong Yang\corref{cor1}}
\author[inst1,inst3]{Yang Xiang\corref{cor2}}
\ead{maxiang@ust.hk}
\author[inst2]{Wenrui Hao\corref{cor2}}
\ead{wxh64@psu.edu}

\affiliation[inst1]{organization={Department of Mathematics},
            addressline={The Hong Kong University of Science and Technology}, 
            state={Clear Water Bay},
            country={Hong Kong Special Administrative Region of China}}

\affiliation[inst2]{organization={Department of Mathematics},
            addressline={The Pennsylvania State University},
            country={USA}}

\affiliation[inst3]{organization={Algorithms of Machine Learning and Autonomous Driving Research Lab},
            addressline={HKUST Shenzhen-Hong Kong Collaborative Innovation Research Institute}, 
            city={Futian},
            state={Shenzhen},
            country={China}}

\begin{abstract}  
Neural network-based approaches have recently shown significant promise in solving partial differential equations (PDEs) in science and engineering, especially in scenarios featuring complex domains or incorporation of empirical data. One advantage of the neural network methods for PDEs lies in its automatic differentiation (AD), which necessitates only the sample points themselves, unlike traditional finite difference (FD) approximations that require nearby local points to compute derivatives. In this paper, we quantitatively demonstrate the advantage of AD in training neural networks. The concept of truncated entropy is introduced to characterize the training property. Specifically, through comprehensive experimental and theoretical analyses conducted on random feature models and two-layer neural networks, we discover that the defined truncated entropy serves as a reliable metric for quantifying the residual loss of random feature models and the training speed of neural networks for both AD and FD methods. Our experimental and theoretical analyses demonstrate that, from a training perspective, AD outperforms FD in solving PDEs.
\end{abstract}



\end{frontmatter}


\section{Introduction}

\begin{figure}[!ht]
    \centering
    \setcounter {subfigure} {0} (a){
    \includegraphics[scale=0.35]{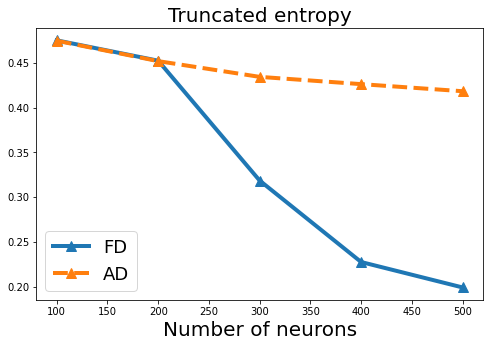}}
    \centering
    \setcounter {subfigure} {0} (b){
    \includegraphics[scale=0.35]{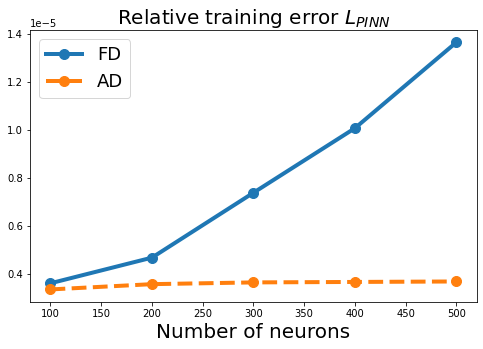}}
      \caption{\textbf{(a):} Truncated entropy. \textbf{(b):} Relative training error $L_{\text{PINN}}$ ($\|\vA \va-\vf\|/\|\vf\|$). They are depicted for both AD and FD methods with varying numbers of neurons in RFM for solving $u_{xx}=f(x)$ with Dirichlet boundary conditions. The exact solution is given by $u(x) = \sin(\pi x)$, where $x\in [-1,1]$. (The number of sample points equals the number of neurons, and the effective cutoff number is $e_A(10^{-12})$). The corresponding singular value of different $M$ and $N$ can be seen in Figure~\ref{Fig.Appendix2.1} in Appendix.}
\label{Residual svd1}
\end{figure}

Neural networks have been extensively applied in solving partial differential equations (PDEs) \cite{raissi2019physics,yu2018deep,han2018solving,siegel2023greedy,hong2021priori,cao2021choose,chen2022bridging,LAN2023112343,chen2025learn,hao2025newton,hao2024multiscale}, particularly in addressing high-dimensional problems, complex domains, and incorporating empirical data. Models like Physics-Informed Neural Networks (PINNs) \cite{raissi2019physics,de2022error,yang2024deeper} and the Deep Ritz method \cite{yu2018deep,lu2021deep} have demonstrated this approach effectively. In solving PDEs using neural networks, the computation of derivatives in the loss functions is crucial. A long-standing debate exists on how to handle differential operators effectively, for which there are primarily two approaches: 
\begin{itemize}
    \item Automatic Differentiation (AD)~\cite{raissi2019physics,de2022error,yang2024deeper,baydin2018automatic,CHIU2022114909}: This method leverages the backpropagation capabilities of neural networks to compute derivatives directly. While AD is precise, it can be memory-intensive and time-consuming, particularly for deep neural networks.
    \item Numerical Differentiation~\cite{grossmann2007numerical, PATEL2022110754,lim2022physics,praditia2021finite,wandel2020learning,wandel2021teaching,gao2021phygeonet,chiu2022can}: Techniques such as the finite difference methods approximate differential operators numerically. Though less accurate and inherently approximate, this approach is easy to implement and computationally less demanding.
\end{itemize}
While some works compare these methods to determine which is superior~\cite{lim2022physics,chiu2022can}, there is limited research focusing on the training aspect. In this paper, we use the finite difference (FD) method as an example of numerical differentiation. We find that there is a significant discrepancy in the training performance between these two differentiation approaches (see Figure \ref{Residual svd1}(b)), with AD outperforming FD. Motivated by this difference, this paper specifically delves into analyzing the training error of these two methods.


In this paper, we compare AD and FD methods for PDEs by training two-layer neural networks and the random feature model (RFM) \cite{chen2022bridging}. Both models follow the same neural network structure (as depicted in Eq. \ref{structure}). The two-layer neural networks (NN) involve training all parameters, whereas the RFM fixes the inner layer parameters randomly and trains only the outer layer parameters. Specifically, the RFM leads to a convex optimization (least squares problem), solving $\vA\va=\vf$, where $\va$ represents the outer layer parameters (as shown in Eq. \ref{structure}). We refer to $\vA$ as the system matrix. In contrast, training the two-layer neural networks involves nonconvex optimization, tackled via gradient descent; the loss dynamics are expressed as $\frac{\partial L}{\partial t}=-\ve^\T \vG \ve$, where $L=\ve^2$ denotes the loss function in PDE solving, and $\vG$ represents the kernel of the gradient descent.

Our analysis is grounded in eigenvalue studies in the training process, shedding light on the distinct behaviors exhibited by AD and FD in solving PDEs using neural network structures. We observe that different differentiation methods influence both $\vA$ and $\vG$, thereby affecting the training process. By analyzing the singular values or eigenvalues of $\vA$ and $\vG$ derived from these differentiation methods, we find that training performance is closely tied to these values. Our theoretical and numerical analyses indicate that AD outperforms FD methods. In RFM, AD achieves a smaller residual error (i.e., training error) (as illustrated in Figure \ref{Residual svd1}(b)). In two-layer neural networks, AD facilitates faster gradient descent dynamics compared to FD, leading to a more rapid decrease in the loss function.

Based on the above phenomena we observe, we provide theoretical and experimental analyses. We find that the singular values of matrices $\vA$ and $\vG$ exhibit similar characteristics in both the AD and FD methods. Both $\vA$ and $\vG$, the large singular values are nearly identical between AD and FD since FD serves as a numerical approximation of AD (Proposition.~\ref{large}). Apart from that, the small singular values of FD are larger than those of AD, primarily due to the numerical differentiation errors on these smaller values (Proposition.~\ref{small}). In the RFM, to avoid the computational error, a truncated singular value approach is necessary to solve $\vA\va=\vf$ by computing a pseudo-inverse of $\mathbf{A}$, where only singular values greater than the truncated value are inverted. Since FD retains more small singular values than AD, this discrepancy leads to relatively larger training errors for FD. These small singular values, while contributing less to approximation error, significantly impact training error, explaining better performance of AD in the RFM. Similarly, in two-layer neural networks, the abundance of small values in $\mathbf{G}$ for FD slows down the training process in gradient descent methods, resulting in slower training compared to AD.

To quantify this observation, we introduce the following two definitions to measure the number of small singular values near the truncation threshold:
\begin{defi}[\textbf{Effective cut-off number}]\label{cutoff}
    Let $\sS$ represent the set of singular values of matrix $\vA$ and $\sigma_{\text{max}}$ is the largest singular value of $\vA$. We define a function $e_{\vA}(a)$ that denotes the count of singular values in set $\sS$ that exceed a given value, denoted by $a \cdot \sigma_{\text{max}}$ ($0\leq a<1$). Mathematically, this can be expressed as:
    \begin{equation}
        e_{\vA}(a) = |\{\sigma \in \sS : \sigma \geq a \cdot \sigma_{\text{max}}\}|, 
    \end{equation}
    where $|\cdot|$ denotes the cardinality of this set.
\end{defi} 

Based on the effective cut-off number, we introduce the  truncated entropy:

\begin{defi}[\textbf{Truncated entropy}]
For a matrix $\vA\in\sR^{m\times n}$ with singular values $\sigma_1\ge\sigma_2\ge\ldots\ge\sigma_{m}$, and a given $a\ge 0$, the truncation entropy of $\vA$ is defined as:
\begin{align}
    H_{\vA}(a) = -\frac{1}{\log e_{\vA}(a)}\sum_{k=1}^{e_{\vA}(a)} p_k \log p_k,
\end{align}
where $p_k=\frac{\sigma_k}{\|\vsigma\|_1}$,for $k=1,2,\ldots, e_{\vA}(a)$, with $\vsigma = [\sigma_1,\sigma_2,\dots,\sigma_{e_{\vA}(a)}]^{T}$ .
\end{defi}

\begin{rmk}
    As defined in Definition 2, if all the singular values are equal, i.e. $\sigma_1 = \sigma_2 =  \dots = \sigma_{e_{\vA}}(a)$,  then $p_k=\frac{\sigma_k}{\|\vsigma\|_1}= \frac{1}{e_{\vA}(a)}$. Therefore $H_{\vA}(a) = -\frac{1}{e_{\vA}(a)} \sum^{e_{\vA}(a)}_{k=1} p_k \log p_k = -\frac{1}{\log e_{\vA}(a)} \sum^{e_{\vA}(a)}_{k=1} \frac{1}{e_{\vA}(a)} \log (\frac{1}{e_{\vA}(a)}) = 1,$ which is also the upper bound for the defined truncated entropy due to Gibbs' inequality. This means that the simplicity of computing the pseudo-inverse for matrices with identical eigenvalues is well-captured by the fact that the truncated entropy achieves its maximum under such circumstances.
\end{rmk}

Note that the defined truncated entropy, which is a normalized version of the spectral entropy~\cite{roy2007effective,frenkel2023understanding} after truncation, can reflect the impact of small singular values after truncation. Theoretical analysis, presented in Theorem \ref{speed}, demonstrates the effectiveness of the defined truncated entropy in quantifying the training speed of gradient descent. Specifically, the larger the value of $H_{\vA}(a)$, the faster the training speed. Empirically, we observe a strong correlation between truncated entropy and training error, as depicted in Figure~\ref{Residual svd1}. As the number of neurons increases, the truncated entropy decreases, leading to an increase in training error. The truncated entropy of AD surpasses that of FD, indicating that AD is able to achieve a lower training error compared to FD.

We delve into a comparative analysis of the performance of automatic differentiation (AD) and finite difference (FD) methods within Physics-Informed Neural Networks (PINN). We present numerical results in higher-dimensional spaces and for higher-order derivative PDEs, to further validate our findings and demonstrate the consistency of results across varying dimensions and complexities.

The remainder of this paper is structured as follows: Section~\ref{sec2 structure} provides a concise overview of neural network methods for solving partial differential equations (PDEs). In section~\ref{PINNsec}, we compare AD and FD, both theoretically and numerically. Section~\ref{RFM} delves into the theoretical relationship between AD and FD with respect to singular values within the framework of the random feature model, through Proposition~\ref{large} and Proposition~\ref{small}. Focusing on a two-layer neural network in section~\ref{NN}, we establish a connection between the defined truncated entropy and training speed, formalized through Theorem~\ref{speed}. Finally, the numerical results on the 2D Poisson equation (Laplacian operator) and biharmonic equation (4th-order differential operator), along with experiments utilizing the 1D Poisson Equation as a case study on a Deep neural network structure, are presented in Section \ref{experiments}.

\section{Neural Network Method for Solving PDEs}\label{sec2 structure}

In this section, we review two types of neural network models: two-layer neural networks and random feature models for solving the PDEs based on the Physics-Informed Neural Networks (PINNs) method \cite{raissi2019physics}.

\subsection{Neural Network Structure}
We focus on the two-layer neural network structure operating in a $d$-dimensional space, where there is only one hidden layer. The model is represented as:
\begin{equation}
\phi(\vx;\boldsymbol{\theta}) = \sum_{j=1}^M a_j \sigma(\vw_j \cdot \vx + b_j),~\vx\in\Omega\label{structure}
\end{equation}
In this neural network, $\boldsymbol{\theta}$ denotes all the parameters $\{(a_j, \vw_j, b_j)\}_{j=1}^M\subset \mathbb{R}\times\mathbb{R}^d\times\mathbb{R}$, all of which are free parameters that need to be learned and $\sigma$ represents the nonlinear activation function. Due to the inherent nonlinearity in this formulation, the learning process becomes nonconvex, posing challenges in optimization.

The random feature model is identical to Eq.~\ref{structure}, except that the inner parameters $\{\vw_j, b_j\}_{j=1}^M$ are randomly chosen and fixed \cite{chen2022bridging}. A common choice is the uniform distribution $\vw_{j} \sim \mathbb{U}\left([-R_{m}, R_{m}]^{d}\right)$ and $b_{j} \sim \mathbb{U}\left([-R_{m}, R_{m}]\right)$, though different distributions can be used. The outer parameters $\{a_j\}_{j=1}^M$ are free and the only ones subject to training. Consequently, the training process can be reduced to a convex optimization problem (least squares problem).

In neural network methods for solving PDEs, the error can be divided into three components: approximation error, generalization error, and training error. The error against the ground truth is bounded by these three errors.
The approximation capabilities of these two models have been extensively studied in the literature. Notable works include \cite{barron1993universal, mhaskar1996neural, lu2021priori, yang2024nearly}, which investigate the approximation ability of neural networks;  the approximation capabilities of the random feature model are studied in \cite{rahimi2008uniform, rahimi2008weighted}. In this paper, we focus on the training error introduced by differential methods, which is distinct from approximation and generalization errors. This provides insights into how to train the loss function when derivative information is included effectively.

\subsection{Solving PDEs Using Neural Network}

For analytical convenience, we illustrate our analysis in the subsequent sections using the Poisson equation as an example; similar analytical techniques can be extended to other types of differential operators. We will extend our results to other equations in the numerical results section. 
Consider the following Poisson equation:
\begin{equation}
\begin{cases}
\Delta u=f & \text { in } \Omega, \\
u =0 & \text { on } \partial \Omega.
\end{cases}
\label{PDE}
\end{equation}
Using a neural network structure, the training loss function of PINNs \cite{raissi2019physics} is defined as:
\begin{equation}
L_\text{PINN}(\boldsymbol{\theta}) := L_\text{F}(\theta) + \lambda L_\text{B}(\theta) =  \sum_{i=1}^N |\Delta\phi(\vx_i;\boldsymbol{\theta}) - f(\vx)|^2 + \lambda \sum_{i=1}^{\widehat{N}} |\phi(\vy_i;\boldsymbol{\theta})|^2.\label{PINN}\end{equation}
Here, $L_\text{F}(\theta)$ represents the residual on PDE equations, while $L_\text{B}(\theta)$ represents the boundary/initial conditions. The variables
$\vx_j \in \Omega$ and $\vy_j \in \partial\Omega$, and $\lambda$ is a constant used to balance the contributions from the domain and boundary terms in the loss function.


\section{AD v.s. FD Methods in Terms of Training}\label{PINNsec}

When solving PDEs using neural networks, the derivatives appear in the loss function, Eq. \ref{PINN}, are mainly handled by two approaches: AD computes derivatives analytically via the chain rule \cite{baydin2018automatic,raissi2019physics,CHIU2022114909}, while FD approximates derivatives numerically based on local points \cite{grossmann2007numerical, PATEL2022110754,lim2022physics,praditia2021finite}. In this section, we compare AD and FD methods based on the loss function of the PINN setup for both the random feature model and two-layer neural networks from the training perspective. Specifically, we investigate the truncated entropy of system matrix $\vA$ in the random feature method, and the training kernel $\vG$ in the neural network. By analyzing these two models, we can evaluate how the differentiation methods influence the training process. For simplicity, we employ the one-dimensional Poisson equation:
\begin{equation}
u_{xx}(x) = f(x), \quad x\in [-1,1]
\label{Eqn.1DPoisson}
\end{equation}
with Dirichlet boundary conditions to illustrate our experimental and theoretical analysis. The exact solution of the equation is represented by $u(x) = \sin(\pi x)$, where $x\in [-1,1]$.

\subsection{AD v.s. FD for Random Feature Models}\label{RFM}
For simplicity, we focus on the differential equation residual loss (since the boundary condition does not involve derivatives)  in Eq.~\ref{PINN}, namely, 
\begin{equation}
L_\text{F}(\va) := \sum_{i=1}^N |\phi_{xx}(x_i;\va) - f(x_i)|^2. \end{equation}
Here, $x_i=-1 + i\cdot h\hbox{~,where~}h=\frac{2}{N}$.
In this context,  since we only need to train the parameters \( \{a_j\}_{j=1}^M \) in the random feature model, we use  \( \phi(x;\va) \) instead of \( \phi(x;\vtheta) \).

To compute the second-order derivatives, the loss function with the AD method can be expressed as:
\begin{equation}
    L_\text{F}^{\text{AD}}(\va) :=\sum_{i=1}^N\left|\sum_{j=1}^M a_j w_{j}^2(\sigma''(w_j\cdot x_i+b_j) -f(x_i)\right|^2 =  \|\vA_{\text{AD}}\cdot\va-\vf\|^2,
\end{equation}where $\vf=[f(x_1),f(x_2),\cdots,f(x_{N})]^T$.
The solution can be computed as \(\va=\vA_{\text{AD}}^{\dagger}\vf
\), where $\vA_{\text{AD}}^{\dagger}=(\vA_{\text{AD}}^\T\vA_{\text{AD}})^{-1}\vA_{\text{AD}}^\T$ is the pseudo-inverse of $\vA_{\text{AD}}$. 

The loss function with the FD method reads as:
\begin{equation}
\begin{aligned}
    L_\text{F}^{\text{FD}}(\va)&=\sum_{i=1}^N\left|\sum_{j=1}^M a_j\frac{\sigma(w_j\cdot x_{i-1}+b_j)+\sigma(w_j\cdot x_{i+1}+b_j)-2\sigma(w_j\cdot x_i+b_j)}{h^2}-f(x_i)\right|^2 \\
    &=  \|\vA_{\text{FD}}\cdot\va-\vf\|^2, 
\end{aligned}
\end{equation} The solution can be computed as 
\(\va=\vA_{\text{FD}}^{\dagger}\vf,\), where $\vA_{\text{FD}}^{\dagger}=(\vA_{\text{FD}}^\T\vA_{\text{FD}})^{-1}\vA_{\text{FD}}^\T$ is the pseudo-inverse of $\vA_{\text{FD}}$. Here $\vA_{\text{AD}}$ and $\vA_{\text{FD}}$ is \begin{equation}
   \vA_{\text{AD}}=\vA_2 \cdot \vD_2,
   \end{equation}
   \begin{equation}
   \vA_{\text{FD}}=\vC_2\cdot \vA_0,
\end{equation} where $\vA_0=(\sigma(w_j\cdot x_i+b_j))_{ij}$, $\vA_2=(\sigma''(w_j\cdot x_i+b_j))_{ij}$,

\begin{equation}
\vC_2 = \frac{1}{h^2}\left[\begin{array}{ccccc}
-2 & 1 & & & \\
1 & -2 & 1 & & \\
& \ddots & \ddots & \ddots & \\
& & 1 & -2 & 1 \\
& & & 1 & -2
\end{array}\right]\in \sR^{N\times N}, \quad \vD_2 =
\begin{bmatrix}
w^2_{1} & & \\
& \ddots & \\
& & w^2_{M}
\end{bmatrix}\in \sR^{M\times M}.
\label{matrix}
\end{equation} 


In this section, we will analyze the training process of these two methods based on the singular values of \( \vA_{\text{AD}} \) or \( \vA_{\text{FD}} \), i.e., the positive square root of eigenvalues of \( \vA_{\text{AD}}^\T\vA_{\text{AD}} \) and \( \vA_{\text{FD}}^\T\vA_{\text{FD}} \). First of all, we want to mention that the large singular values are close for the two matrices since when \( h \) is small, the gap between the two matrices is small, i.e.,
\begin{equation}
    \vA_{\text{AD}} = \vA_{\text{FD}} + h^2 \vE,
\end{equation}
where $\vE$ is the numerical differentiation errors matrix. Hence, the difference in eigenvalues between the two matrices is  \(\mathcal{O}(h^2)\), which will affect the small singular values a lot but not the large singular values. We summarize in the following propositions, which we are
\begin{prop}\label{large}
    Denote the largest and smallest eigenvalue \( \vE^\T\vA_{\text{FD}}+\vA_{\text{FD}}^\T\vE+h^2 \vE^\T\vE \) as \( \bar{\lambda}, 
 \underline{\lambda} \), we have that
    \begin{equation}
        \lambda_{\max}(\vA_{\text{FD}}^\T\vA_{\text{FD}})+h^2\underline{\lambda}\leq\lambda_{\max}(\vA_{\text{AD}}^\T\vA_{\text{AD}})\leq \lambda_{\max}(\vA_{\text{FD}}^\T\vA_{\text{FD}})+h^2 \bar{\lambda},
    \end{equation}where $\lambda_{\max}(\vA)$ is denoted as the largest eigenvalue of $\vA$.
\end{prop}
\begin{proof}
    First of all, we prove 
\[
\lambda_{\max}(\vA_{\text{AD}}^\T\vA_{\text{AD}}) \leq \lambda_{\max}(\vA_{\text{FD}}^\T\vA_{\text{FD}}) + h^2 \bar{\lambda}.
\]
Since  
\begin{equation}  
    \vA_{\text{AD}}^\T\vA_{\text{AD}} = \vA_{\text{FD}}^\T\vA_{\text{FD}} + h^2 (\vE^\T\vA_{\text{FD}} + \vA_{\text{FD}}^\T\vE) + h^4 \vE^\T\vE,  
\end{equation}  
it follows that  
\begin{equation}  
    [\lambda_{\max}(\vA_{\text{FD}}^\T\vA_{\text{FD}}) + h^2 \bar{\lambda}]\cdot \vI - \vA_{\text{AD}}^\T\vA_{\text{AD}}  
\end{equation}  
is a positive definite matrix. This result is due to Weyl's inequalities, which state that for any symmetric matrices \(\vA\) and \(\vB\),  
\begin{equation}  
    \lambda_{\max}(\vA + \vB) \leq \lambda_{\max}(\vA) + \lambda_{\max}(\vB).  
\end{equation}
 For any eigenvalue of $\vA_{\text{AD}}^\T\vA_{\text{AD}}$, $\lambda$, $\vA_{\text{AD}}^\T\vA_{\text{AD}} \vv = \lambda \vv$, we have 
\begin{equation}
    [\lambda_{\max}(\vA_{\text{FD}}^\T\vA_{\text{FD}}) + h^2 \bar{\lambda}]\cdot \vI \cdot \vv - \vA_{\text{AD}}^\T\vA_{\text{AD}}\vv = (\lambda_{\max}(\vA_{\text{FD}}^\T\vA_{\text{FD}}) + h^2 \bar{\lambda} - \lambda)\vv.
\end{equation} 
Hence 
\[
\lambda_{\max}(\vA_{\text{FD}}^\T\vA_{\text{FD}}) + h^2 \bar{\lambda} - \lambda \geq 0,
\]
due to the arbitrary choice of $\lambda$, we obtain 
\[
\lambda_{\max}(\vA_{\text{AD}}^\T\vA_{\text{AD}}) \leq \lambda_{\max}(\vA_{\text{FD}}^\T\vA_{\text{FD}}) + h^2 \bar{\lambda}.
\]
Now we prove the next part, we have 
\begin{equation}
    (\lambda_{\max}(\vA_{\text{AD}}^\T\vA_{\text{AD}}) -h^2\underline{\lambda})\cdot \vI- \vA_{\text{FD}}^\T\vA_{\text{FD}}
\end{equation}
is a positive definite matrix. For any eigenvalue of $\vA_{\text{FD}}^\T\vA_{\text{FD}}$, $\lambda$, $\vA_{\text{FD}}^\T\vA_{\text{FD}} \vv = \lambda \vv$, we have 
\begin{equation}
    [(\lambda_{\max}(\vA_{\text{AD}}^\T\vA_{\text{AD}}) -h^2\underline{\lambda}) \cdot \vI - \vA_{\text{FD}}^\T\vA_{\text{FD}}]\vv = (\lambda_{\max}(\vA_{\text{AD}}^\T\vA_{\text{AD}})- h^2\underline{\lambda}-\lambda)\vv.
\end{equation} 
Hence 
\[
\lambda_{\max}(\vA_{\text{AD}}^\T\vA_{\text{AD}}) - h^2\underline{\lambda}-\lambda \geq 0,
\]
due to the arbitrary choice of $\lambda$, we obtain 
\[
\lambda_{\max}(\vA_{\text{FD}}^\T\vA_{\text{FD}})+ h^2\underline{\lambda}\leq \lambda_{\max}(\vA_{\text{AD}}^\T\vA_{\text{AD}}).
\]
\end{proof}

Based on Proposition \ref{large}, we have $\sigma_{\max}(\vA_{\text{AD}})\approx \sigma_{\max}(\vA_{\text{FD}})$ when $h\ll1$. For the small singular value, based on the following Proposition~\ref{small}, we know that the small singular value, i.e., the singular value in \(\fO(h)\) level, \( \vA_{\text{AD}} \), should be smaller than \( \vA_{\text{FD}} \). 

\begin{prop}\label{small}
    Suppose $\vD_2$ is invertible, $\sigma_{\min}(\vA_2)\not=0$ and \begin{equation} \label{assimplink} C:=\frac{\sigma_{\min}(\vA_0)}{\sigma_{\min}(\vA_2) } \geq \frac{1}{\lambda_{\min}(\vC_2)\lambda_{\min}(\vD_2^{-1})}, \end{equation}then we have
    \begin{equation}
       \sigma_{\min}(\vA_{\text{FD}}) \geq \sigma_{\min}(\vA_{\text{AD}}),
    \end{equation}where $\sigma_{\min}(\vA)$ is denoted as the smallest singular value of $\vA$.
\end{prop} 

\begin{proof}
First of all, we show that \begin{equation}
        \sigma_{\min}(\vA_{\text{FD}}) \geq \sigma_{\min}(\vA_0)\lambda_{\min}(\vC_2), \quad \sigma_{\min}(\vA_{2}) \geq \sigma_{\min}(\vA_{\text{AD}})\sigma_{\min}(\vD_2^{-1}).
    \end{equation}
   We know that for the symmetric positive definite matrices \( \vA \) and \( \vB \), \( \lambda_{\max} \) is a kind of sub-multiplicative norm, i.e., 
\[
\lambda_{\max}(\vA\vB) \leq \lambda_{\max}(\vA)\lambda_{\max}(\vB).
\]
By taking the inverse, we have 
\[
\lambda_{\min}(\vA\vB) \geq \lambda_{\min}(\vA)\lambda_{\min}(\vB).
\]
Therefore, we have 
\begin{align*}
    \lambda_{\min}(\vA_{\text{FD}}^\T\vA_{\text{FD}}) = \lambda_{\min}(\vA^\T_0\vC^\T\vC_2\vA_0) 
    \geq \lambda_{\min}(\vA_0^\T\vA_0)\lambda^2_{\min}(\vC_2).
\end{align*}
Thus,
\[
\sigma_{\min}(\vA_{\text{FD}}) \geq \sigma_{\min}(\vA_0)\lambda_{\min}(\vC_2).
\]
The other inequality is proved in the same way.

Based on definition of $C$, we have 
\[
\sigma_{\min}(\vA_{\text{FD}}) \geq C\lambda_{\min}(\vC_2)\lambda_{\min}(\vD_2^{-1}) \sigma_{\min}(\vA_{\text{AD}}).
\]
Due to \( C \geq \frac{1}{\lambda_{\min}(\vC_2)\lambda_{\min}(\vD_2^{-1})} \), we have \( \sigma_{\min}(\vA_{\text{FD}}) \geq \sigma_{\min}(\vA_{\text{AD}}) \).
\end{proof}

\begin{rmk}
    In the above proposition, we assume (\ref{assimplink}), which is reasonable if we omit the effect of the activation function. For smooth activation functions such as \( \sin(x) \) and \( \tanh(x) \), the eigenvalues of the derivative matrix are of the same order of magnitude as those of the original matrix. In other words, we have \( \sigma_{\min}(\vA_0) = C\sigma_{\min}(\vA_2) \), where $C$ is close to 1. Since \( \lambda_{\min}(\vC_2) = N^2 (\cos \frac{\pi}{N+1}-1) > 1 \)\cite{morton2005numerical}, and \( \lambda_{\min}(\vD_2^{-1}) > 1 \) if we randomly choose $w_{i} \sim \mathbb{U}\left((-R_m, R_m)^{d}\right)$, Eq.~(\ref{assimplink}) will hold.
\end{rmk}

The two propositions above provide theoretical analysis into the distinctions between the singular values of $\vA_{\text{AD}}$ and $\vA_{\text{FD}}$. Proposition~\ref{large}  indicates the similarity of the large singular values in $\vA_{\text{AD}}$ and $\vA_{\text{FD}}$, while Proposition~\ref{small} states that small singular values in $\vA_{\text{FD}}$ are larger than those in $\vA_{\text{AD}}$. 

In this paper, we use the Poisson equation in the theoretical analysis for simplicity and ease of notation. The proposed framework can be extended to other PDEs, as the core analysis remains applicable across different equations. The generality of Proposition 1 and Proposition 2 in the paper can be understood as follows. 

For Proposition 1, since FD is an approximation of AD, the large eigenvalues in the kernel of the training dynamics are similar for both methods. This property holds universally across PDEs. For Proposition 2, as long as the condition $C:=\frac{\sigma_{\min}(\vA_0)}{\sigma_{\min}(\vA_2) } \geq \frac{1}{\lambda_{\min}(\vC_2)\lambda_{\min}(\vD_2^{-1})}$ is satisfied (where these matrices are defined in Eqs. (10)-(12)), we can establish the conclusion of Proposition 2 that 
$\sigma_{\min}(\vA_{\text{FD}}) \geq \sigma_{\min}(\vA_{\text{AD}})$.  
This condition can be satisfied for different PDEs. 
Roughly speaking, consider the $k$-th order derivative $\frac{d^k}{dx^k}$, and set $\vA_0 = (\sigma(w_j \cdot x_i + b_j))_{ij}$ and $\vA_k = (\sigma^{(k)}(w_j \cdot x_i + b_j))_{ij}$. The activation function $\sigma$ commonly used for neural networks solving PDEs is $\sin(x)$, which leads to $\frac{\sigma_{\min}(\vA_0)}{\sigma_{\min}(\vA_k)}$ being approximately equal to 1. For $\vC_k$, the discrete matrix for the derivative operator in general tends to be large, meaning that $\lambda_{\min}(\vC_k) > 1$.  For $\vD_k$, 
in our analysis, we adopt the framework of the random feature model, where $w_i \sim \mathbb{U}\left((-R_m, R_m)^{d}\right)$, which leads to $\lambda_{\min}(\vD_k^{-1}) > 1$. We have 
 $\lambda_{\min}(\vC_k) \lambda_{\min}(\vD_k^{-1}) \gg 1$. Thus,  the condition $C := \frac{\sigma_{\min}(\vA_0)}{\sigma_{\min}(\vA_k)} \geq \frac{1}{\lambda_{\min}(\vC_k) \lambda_{\min}(\vD_k^{-1})}$ can also hold for differential operators of different orders. For other domain sizes, we can introduce the nondimensional variable $\xi = 2\frac{x - a}{(b - a)} - 1$, under which, for example, the second derivative transforms as $\frac{d^2}{d\xi^2} = \frac{(b - a)^2}{4}\frac{d^2}{dx^2}$. The analysis remains unchanged. 

 For example, in the case of the biharmonic equation (the fourth-order derivative), we have \( \lambda_{\min}(\vC_4)\approx \frac{N^4}{4}(\cos(\frac{\pi}{N+1})-1)^2 > 1 \)(where $\vC_4$ is the discrete matrix for the fourth-order derivative), and \( \lambda_{\min}(\vD_4^{-1}) > 1 \) when randomly choosing $w_{i} \sim \mathbb{U}\left((-R_m, R_m)^{d}\right)$. This ensures that the condition and thus the conclusion of Proposition 2 holds, meaning that the smaller eigenvalues of FD are consistently larger than those of AD for the biharmonic equation as well.

Moreover, as demonstrated in numerical experiments in Section 4, the findings remain valid across different PDEs.

\begin{figure}[!ht]
    \centering   
    \setcounter {subfigure} {0} (a){\includegraphics[scale=0.44]{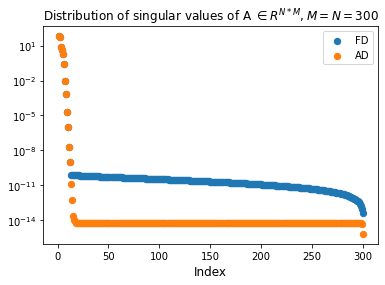}}
    \setcounter {subfigure} {0} (b){\includegraphics[scale=0.44]{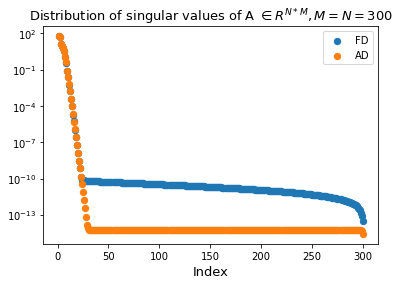}}
    \setcounter {subfigure} {0} (c){\includegraphics[scale=0.43]{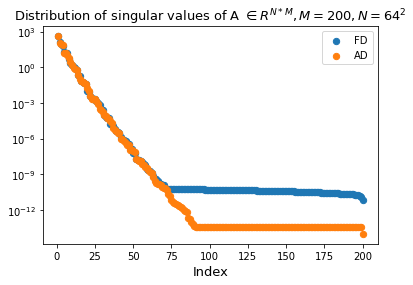}}
    \setcounter {subfigure} {0} (d){\includegraphics[scale=0.43]{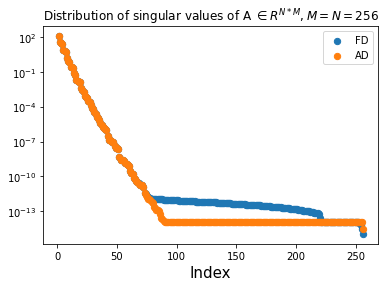}}
    \caption{Distribution of singular values for $\vA_{\text{FD}}$ and $\vA_{\text{AD}}$ for solving Poisson equation using random feature method with vary Dimensions $d$, activation functions $\sigma(x)$, 
number of sample points $N$ and number of neurons $M$. \textbf{(a):} $d=1$, $\sigma(x)=\sin(x)$ ,$M=N=100$. \textbf{(b):} $d=1$, $\sigma(x)=\tanh(x)$, $M=N=300$. \textbf{(c):} $d=2$, $\sigma(x) = \sin(x)$, $M=200$, $N=64\times64$. \textbf{(d):} $d=2$, $\sigma(x) = \sin(x)$, $M=N=16\times16$.}
\label{smallbig}
\end{figure}

To validate our analysis, we conduct experiments using the random feature method, where we varied the activation functions and dimensions to solve the Poisson equation. Here we choose  $w_{j}$ and $b_j$ from uniform distribution $\mathbb{U}\left([-1, 1]\right)$ . As depicted in Figure~\ref{smallbig}, we observe that the large singular values of \( \vA_{\text{FD}} \) and \( \vA_{\text{AD}} \) are almost equal, while the small eigenvalue of \( \vA_{\text{FD}} \) is larger than \( \vA_{\text{AD}} \), confirming the analysis presented earlier.

The difference between the AD and FD methods in solving PDEs using random feature models can be elucidated by analyzing the different characteristics of the singular values of $\boldsymbol{A}_\text{AD}$ and $\boldsymbol{A}_\text{FD}$. Minimizing the training loss $L_{F}^\text{AD}(\boldsymbol{a})$ and $L_{F}^\text{FD}(\boldsymbol{a})$ can be reformulated as solving the linear systems $\boldsymbol{A}_{\text{AD}} \boldsymbol{a} = \boldsymbol{f}$ and $\boldsymbol{A}_{\text{FD}}\boldsymbol{a} = \boldsymbol{f}$, respectively. 
To solve $\boldsymbol{A}\boldsymbol{a} = \boldsymbol{f}$, it is often necessary to compute the pseudo-inverse of matrix $\vA$. However, if the non-zero singular values are too small, the computation error becomes large.
To mitigate computation errors, it is common practice to perform truncation based on singular values of matrix $\vA$. Here, we utilize the truncated SVD method to analyze the influence of eigenvalues on the training error. We denote the singular values of $\vA_{\text{AD}}$ as $\{\sigma_i\}_{i=1}^M$ and those of $\vA_{\text{FD}}$ as $\{\widehat{\sigma}_i\}_{i=1}^M$. The SVD decomposition of $\vA_{\text{AD}}$ and $\vA_{\text{FD}}$ can be expressed as follows:\begin{align}
    \vA_{\text{AD}} = \sum_{i=1}^{P} \sigma_i u_i v_i^\T  +\sum_{i=P+1}^{M} \sigma_i u_i v_i^\T,~
    \vA_{\text{FD}} = \sum_{i=1}^{\widehat{P}} \widehat{\sigma}_i \widehat{u}_i\widehat{v}_i^\T  + \sum_{i=\widehat{P}+1}^{M} \widehat{\sigma}_i \widehat{u}_i \widehat{v}_i^\T,
\end{align}
where $P$ and $\hat{P}$ are the truncated positions. Then the truncated matrix can be expressed as
$$
\widetilde{\vA}_{\text{AD}} = \sum_{i=1}^{P} \sigma_i u_i v_i^\T,~ \widetilde{\vA}_{\text{FD}} = \sum_{i=1}^{\hat{P}} \widehat{\sigma}_i \widehat{u}_i \widehat{v}_i^\T.
$$
By direct calculation, the truncation error can be expressed as 
\begin{align*}
\epsilon_{\text{AD}} := \left\|\vA_{\text{AD}}-\widetilde{\vA}_{\text{AD}}\right\|^{2} = \sum_{i=P+1}^{M} \sigma_i^2,~
\epsilon_{\text{FD}} = \left\|\vA_{\text{FD}}-\widetilde{\vA}_{\text{FD}}\right\|^{2} = \sum_{i=\hat{P}+1}^{M} \widehat{\sigma}_i^2.
\end{align*} Thus the training error can be formulated as
\begin{equation}
\begin{aligned}
        \|\vA_{\text{AD}}\cdot\va-\vf\|^2 \leq \|\widetilde{\vA}_{\text{AD}}\cdot\va-\vf\|^2 +  \epsilon_{\text{AD}}\|\va\|,\\  
        \|\vA_{\text{FD}}\cdot\va-\vf\|^2  \leq \|\widetilde{\vA}_{\text{FD}}\cdot\va-\vf\|^2 +  \epsilon_{\text{FD}}\|\va\|.
\end{aligned}
\label{SVDerror}
\end{equation}

\begin{figure}[!ht]
    \setcounter {subfigure} {0} (a){
    \includegraphics[scale=0.35]{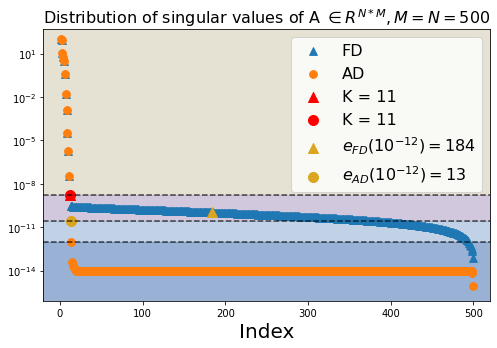}}
    \setcounter {subfigure} {0} (b){
    \includegraphics[scale=0.35]{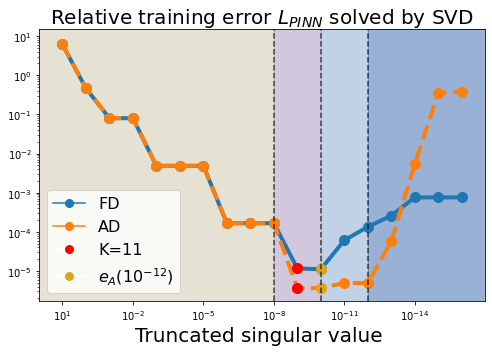}}
    \caption{\textbf{(a):} The distribution of singular values alongside their respective truncation positions. Green dots represent truncated positions based on $e_{\vA}(10^{-12})$, with \textit{truncated etntropy} values of $H_{\vA_\text{FD}}(10^{-12}) = 0.1995$ and $H_{\vA_\text{AD}}(10^{-12}) = 0.4183$. \textbf{(b):} Relative training error of $L_\text{PINN}$, denoted as $\frac{\|\vA \va -\vf\|}{\|\vf\|}$, obtained using the truncated singular value decomposition (SVD) method.  The horizontal axis represents the truncation position determined by the effective cut-off number. The dashed lines differentiate between different cases based on the truncation position in both \textbf{(a)} and \textbf{(b)}.}
    \label{Fig.cutoff Residual}
\end{figure}



The numerical results are shown in Figure~\ref{Fig.cutoff Residual}. We determine the truncated positions based on $\textit{effective cut-off number}$, i.e., $P = e_{\vA_{\text{AD}}}(a)$ and $\hat{P} = e_{\vA_{\text{FD}}}(a)$. We can divide the truncated position into four cases based on the relationship between training error and the truncated position. In \textbf{Case \uppercase\expandafter{\romannumeral1 }}, the truncated position $P=\hat{P} < K$, where $K$ represents the position of the last similar singular value shared by $\vA_{\text{AD}}$ and $\vA_{\text{FD}}$, as indicated by the red dot in Figure~\ref{Fig.cutoff Residual}(a). The training errors for AD and FD are similar in this case because the training error shown in Eq.~\ref{SVDerror} is dominated by $\epsilon_{\text{AD}} \approx \sum_{i=P+1}^{K}\sigma_i^2 \approx\epsilon_{\text{FD}} \approx \sum_{i=P+1}^{K} \widehat{\sigma}_i^2$ in Eq.~\ref{SVDerror}.  Therefore, in this case, the training errors of AD and FD are similar. In \textbf{Case \uppercase\expandafter{\romannumeral2}}, the truncated position $P \geq K$. The training error is dominated by $\epsilon_{\text{AD}} \approx \sum_{i=P+1}^{M}\sigma_i^2 < \epsilon_{\text{FD}} \approx \sum_{i=P+1}^{M} \widehat{\sigma}_i^2$. This is because the small eigenvalues of FD are larger than those of AD. As shown in the Figure~\ref{Fig.cutoff Residual}(b), the training error of AD is smaller than FD.

In \textbf{Case III} and \textbf{Case IV}, we observe that increasing the truncation position $P$ leads to an increase in training error. In this case, the training error shown in Eq.~\ref{SVDerror} is dominated by $\|\widetilde{\vA}_{\text{AD}}\cdot\va-\vf\|^2$ and $\|\widetilde{\vA}_{\text{FD}}\cdot\va-\vf\|^2$ caused by computation errors that arise when computing the pseudo-inverse of the truncated matrix. As shown in Figure~\ref{Fig.cutoff Residual}(b) if $P>e_{\vA}(10^{-12})$, it would lead to computational errors in computing the pseudo-inverse of $\vA_{\text{AD}}$ and $\vA_{\text{FD}}$, consequently causing an increase in the training error. 

If we consider the case where significant computational errors are not introduced in the SVD method, the optimal truncation position is $e_{A}(10^{-12})$. Therefore, in the SVD approach, the hyperparameter for the effective cut-off number, $a$, can be chosen as $10^{-12}$, i.e., $e_{A}(10^{-12})$. As shown in Figure~\ref{Fig.cutoff Residual}(a), when the truncated position is $P = e_{A}(10^{-12})$, since the small singular values in $\vA_{\text{FD}}$ are larger than those in $\vA_{\text{AD}}$, more small eigenvalues are retained in $\widetilde{\vA}_{\text{FD}}$. By calculating the corresponding truncated entropy we have $H_{\vA_\text{AD}}(10^{-12}) = 0.4183 > H_{\vA_\text{FD}}(10^{-12}) = 0.1995$. The corresponding lower training error in AD compared to FD indicates the effectiveness of truncated entropy.

The variation of training error with the truncated position in the truncated SVD method can provide insights for training in the following neural network case. 

\subsection{AD v.s. FD for Two-layer Neural Networks}\label{NN}
 Unlike random feature models, for neural networks, the parameters in the activation functions are free, and we cannot directly solve the optimization problem using methods like $\vA \va=\vf$, as this problem is non-convex. To tackle such optimization problems, there are various methods available, including gradient descent \cite{ruder2016overview}, stochastic gradient descent \cite{zinkevich2010parallelized}, Adam method \cite{kingma2014adam}, Gauss-Newton method \cite{hao2023gauss}, preconditioned methods \cite{he2024nltgcr}, homotopy methods \cite{yang2023homotopy}, and others. These methods are all gradient descent-based methods. Here our analysis is based on the gradient descent method. 

Due to the structure of neural networks, there are two types of parameters: one is outside the activation functions, i.e., $\{a_j\}_{j=1}^M$, and the other is inside the activation function, $\{w_j,b_j\}_{j=1}^M$ (See Eq.~\ref{structure}). For parameters $\{a_j\}_{j=1}^M$, the gradient descent can be expressed as:
\begin{equation}
\frac{\D \va}{\D t}=-2\vA^\T_\kappa(\Delta_\kappa\vphi(\vx;\vtheta) - f(\vx)),
\end{equation}
where $\vA_\kappa=\vA_{\text{AD}}$ for AD method, $\vA_\kappa=\vA_{\text{FD}}$ for FD method and $\Delta_\kappa$ indicates different approaches to dealing with the Laplacian operator ($\Delta_{\text{AD}}$ for AD and $\Delta_{\text{FD}}$ for FD).   For the parameter $\{w_i,b_i\}_{i=1}^M$, we only consider $w_i$ since $b_i$ can be merged into $w_i$ by setting $\vx=(x,1)$. The gradient descent of AD methods can be expressed as:
\begin{equation}
\frac{\D \vw}{\D t}=-2(\vX\cdot\vA_3\cdot \vD_3\cdot \bar{\vA}+2\vA_2\cdot \vD_1\cdot \bar{\vA})^\T\cdot(\Delta_\kappa\vphi(\vx;\vtheta) - f(\vx)),
\end{equation}
where $\vA_3=(\sigma^{(3)}(w_j\cdot x_i+b_j))_{ij}$, $\bar{\vA}=\operatorname{diag}(a_1,\ldots,a_M)$ and $\vX=\operatorname{diag}(x_1,\ldots,x_N)$.
For gradient descent in FD method, it can be expressed as:
\begin{equation}
\frac{\D \vw}{\D t}=-2(\vC_2\cdot\vX\cdot\vA_1\cdot \bar{\vA})^\T(\Delta_\kappa\vphi(\vx;\vtheta) - f(\vx)).
\end{equation} For the loss function $L_\text{F}(\vtheta) := \sum_{i=1}^N |\Delta_\kappa\phi(x_i;\vtheta) - f(x_i)|^2$. The gradient descent dynamics can be expressed as:
\begin{equation}
\begin{aligned}
\frac{\D L_\text{F}(\vtheta)}{\D t} &= \nabla_{\va}L_\text{F}(\vtheta)\frac{\D \va}{\D t} + \nabla_{\vw}L_\text{F}(\vtheta)\frac{\D \vw}{\D t} \\&=-2(\Delta_\kappa\vphi(\vx;\vtheta) - f(\vx))^\T \vG_\kappa (\Delta_\kappa\vphi(\vx;\vtheta) - f(\vx)),
\end{aligned}
\end{equation}where \begin{align}\vG_\text{AD}&=\vA_\text{AD}\vA_\text{AD}^\T+(\vX\cdot\vA_3\cdot \vD_3\cdot \bar{\vA}+2\vA_2\cdot \vD_1\cdot \bar{\vA})(\vX\cdot\vA_3\cdot \vD_3\cdot \bar{\vA}+2\vA_2\cdot \vD_1\cdot \bar{\vA})^\T,\notag\\\vG_\text{FD} &=\vA_\text{FD}\vA_\text{FD}^\T+(\vC_2\cdot\vX\cdot\vA_1\cdot \bar{\vA})(\vC_2\cdot\vX\cdot\vA_1\cdot \bar{\vA})^\T.\end{align}

\begin{figure}[!ht]
    \centering
    \setcounter {subfigure} 0(a){
    \includegraphics[scale=0.5]{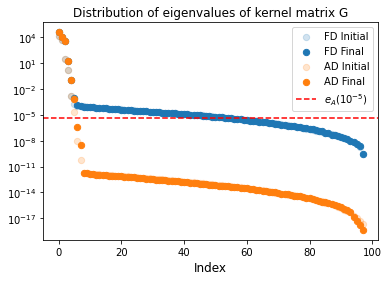}} \\
    \setcounter {subfigure} 0(b){
    \includegraphics[scale=0.35]{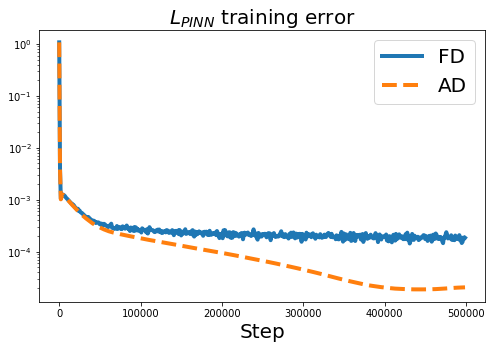}}
    \setcounter {subfigure} 0(c){
    \includegraphics[scale=0.35]{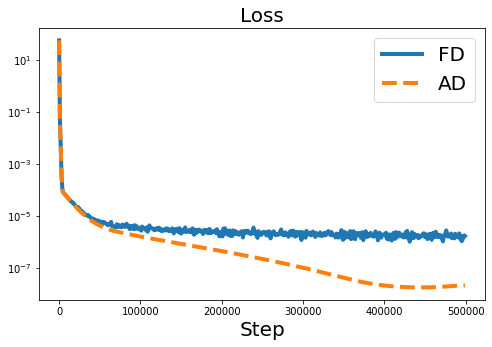}}
    \caption{\textbf{(a):} Distribution of eigenvalues of kernel matrix $G$ with the red dashed line indicating the approximate convergence position $e_{\vG_k}(10^{-5})$ at the end of training. \textit{Truncated entropy} values are $H_{\vG_\text{FD}}(10^{-5}) = 0.5785$ and $H_{\vG_\text{AD}}(10^{-5}) = 0.2606$. \textbf{(b):} Training curve of relative training error $\frac{L_F}{\|f\|}$. \textbf{(c):} Training curve of the loss function $L_{\text{PINN}}$. (Similar performance observed for $L_{F}$ and $L_{\text{PINN}}$ due to the boundary conditions being treated as an identity operator.) More training curves of training error and their corresponding curves of L2 relative error can be viewed in Figures~\ref{Fig.lcurverelativecurve} in appendix. }
    \label{PINNGD}
\end{figure}

As depicted in Figure~\ref{PINNGD}(a), the eigenvalues of the kernel $\vG$ in the gradient descent process for learning $\vtheta$ exhibit similarities to those of system matrix $\vA$ observed in the random feature model case. This is expected, given that the kernels $\vG_{\text{FD}}$ tend to converge to $\vG_{\text{AD}}$ as $h$ diminishes, reducing the difference between the large eigenvalues of $\vG_{\text{FD}}$ and $\vG_{\text{AD}}$ (Proposition \ref{large}). Additionally, $\vC_2$ in Eq.~\ref{matrix} serves to enlarge the small eigenvalues of the kernel in the finite difference (FD) method, making them greater than those in the automatic differentiation (AD) methods (Proposition \ref{small}). A similar analysis applies to $\vA_\kappa$. While the kernel undergoes changes during training the two-layer neural networks, unlike in the random feature model, the eigenvalue relationship between AD and FD remains valid throughout training, as illustrated in Figure~\ref{PINNGD}(a). 
Based on Figure~\ref{PINNGD}(b), in the beginning of the gradient descent, the performances of AD and FD methods are similar, which is due to the dominant role played by the similar large eigenvalues. After several iteration steps in gradient descent, the speed of gradient descent in FD is slower than that of AD. This is due to the fact that the small eigenvalue direction begins to play an important role in the training, and we can explain this theoretically as follows. 

For the vector $\vr_{\text{AD}} := \Delta_\text{AD}\vphi(\vx;\vtheta) - f(\vx)$, we perform the eigenvalue decomposition of $\vA^{\T}_{\text{AD}}\vA_{\text{AD}}$, denoted as
\(
   \vr_{\text{AD}} = \sum_{i=1}^M \vr_{i,\text{AD}}
\)
such that $\vA_{\text{AD}}\vA_{\text{AD}}^{\T}\vr_{i,\text{AD}} = \lambda_{i,\text{AD}} \vr_{i,\text{AD}}$ for $\lambda_{1,\text{AD}} \ge \lambda_{2,\text{AD}} \ge \ldots \ge \lambda_{M,\text{AD}}$. Similarly, for finite difference method, we have
\(
    \vr_{\text{FD}} := \Delta_\text{FD}\vphi(\vx;\vtheta) - f(\vx) = \sum_{i=1}^M \vr_{i,\text{FD}}
\)
such that $\vA_{\text{FD}}\vA_{\text{FD}}^{\T}\vr_{i,\text{FD}} = \lambda_{i,\text{FD}} \vr_{i,\text{FD}}$ for $\lambda_{1,\text{FD}} \ge \lambda_{2,\text{FD}} \ge \ldots \ge \lambda_{M,\text{FD}}$. Therefore, the gradient descent for two methods can be written as
\begin{equation}
     \frac{\D L_{\kappa}(\vtheta)}{\D t} = -2 \sum_{i=1}^M \lambda_{i,\kappa} \vr_{i,\kappa}^2.
\label{Eqn.trainDynamics}
\end{equation}
For eigenvalues smaller than $a  \sqrt{\lambda_1}$ ($a$ is the hyperparameter in Definition 1), the training error can be considered negligible. This is similar to the truncated SVD in RFM, as shown in Figure~\ref{Fig.cutoff Residual}(b), the singular values $\sigma<10^{-12}  \sigma_1$ can be truncated, given that the training error will not decrease if we increase the truncation position. Due to the similar performance of $\vA_\kappa$ and $\vG_\kappa$, we assume the error in this part to be $0$ in the gradient descent.  That is $\vr_{i,\text{FD}} \approx 0$ for $i > e_{\vG_\text{FD}}(a)$ and $\vr_{i,\text{AD}} \approx 0$ for $i > e_{\vG_\text{AD}}(a)$ throughout the entire learning process. This means we disregard these eigenvalues when determining loss convergence speed. Since large eigenvalues (for $i>e_{\vG_{\kappa}(b)}$, where $b>a$.) converge fast, we consider the stage when the convergence speed is dominated by gradient descent of the kept small eigenvalues, i.e., \begin{equation}
     \frac{\D L_{\kappa}(\vtheta)}{\D t} = -2 \sum_{i=e_{\vG_{\kappa}(b)}}^{e_{\vG_{\kappa}(a)}} \lambda_{i,\kappa} \vr_{i,\kappa}^2.
\end{equation} The convergence speed of the above stage is stated in the following theorem:

\begin{thm}\label{speed}
    Suppose there exist a constant $a,b, (b>a),t_*$ such that $\vr_{i,\text{FD}} = 0$ for $i > e_{\vG_\text{FD}}(a),i < e_{\vG_\text{FD}}(b)$, and $\vr_{i,\text{AD}} = 0$ for $i > e_{\vG_\text{AD}}(a),~i < e_{\vG_\text{AD}}(b)$ for all $t \ge t_*$. For any $T \ge t_*$, we have that
    \begin{align}
        &\exp\left(-2 \zeta_{\kappa}(T) \cdot \max_{t \in [t_*,T]} \frac{\sum_{i=e_{\vG_\kappa}(b)}^{e_{\vG_\kappa}(a)}\lambda_{i,\kappa}}{e_{\vG_\kappa}(a)-e_{\vG_\kappa}(b)} (t-t_*)\right) \cdot L_{\kappa}(\vtheta)[t_*]\notag\\\le& L_{\kappa}(\vtheta)[t] \le \exp\left(-2 \eta_{\kappa}(T) \cdot \min_{t \in [t_*,T]} \frac{\sum_{i=e_{\vG_\kappa}(b)}^{e_{\vG_\kappa}(a)}\lambda_{i,\kappa}}{e_{\vG_\kappa}(a)-e_{\vG_\kappa}(b)} (t-t_*)\right) \cdot L_{\kappa}(\vtheta)[t_*],
    \end{align}
    for all $t \in [t_*,T]$, where $\kappa=\text{AD},~\text{FD}$, and
    \begin{equation}
    \eta_{\kappa}(T) = \min_{t \in [t_*,T]}\frac{\min\{\vr_{i,\kappa}^2\}_{i=e_{\vG_\kappa}(b)}^{e_{\vG_\kappa}(a)}}{\max\{\vr_{i,\kappa}^2\}_{i=e_{\vG_\kappa}(b)}^{e_{\vG_\kappa}(a)}},~\zeta_{\kappa}(T):=\max_{t \in [t_*,T]}\frac{\max\{\vr_{i,\kappa}^2\}_{i=e_{\vG_\kappa}(b)}^{e_{\vG_\kappa}(a)}}{\min\{\vr_{i,\kappa}^2\}_{i=e_{\vG_\kappa}(b)}^{e_{\vG_\kappa}(a)}}.
    \label{Constant}
    \end{equation}
\end{thm}

\begin{proof}
Due to \( \frac{\D L_{\kappa}(\vtheta)}{\D t}=-2 \sum_{i=1}^M\lambda_{i,\kappa} \vr_{i,\kappa}^2\) in Eq.~\ref{Eqn.trainDynamics} and  $\vr_{i,\kappa}= 0$ for $i> e_{\vG_\kappa}(a),~i< e_{\vG_\kappa}(b)$, we have that \begin{align}
    \frac{\D L_{\kappa}(\vtheta)}{\D t}&=-2 \sum_{i=e_{\vG_\kappa}(b)}^{e_{\vG_\kappa}(a)}\lambda_{i,\kappa} \vr_{i,\kappa}^2\notag\\&= \frac{-2 \sum_{i=e_{\vG_\kappa}(b)}^{e_{\vG_\kappa}(a)}\lambda_{i,\kappa} \vr_{i,\kappa}^2}{ \sum_{i=e_{\vG_\kappa}(b)}^{e_{\vG_\kappa}(a)}\vr_{i,\kappa}^2}L_{\kappa}(\vtheta)\notag\\&\le-2 \frac{\min\{\vr_{i,\kappa}^2\}_{i=e_{\vG_\kappa}(b)}^{e_{\vG_\kappa}(a)}}{\max\{\vr_{i,\kappa}^2\}_{i=e_{\vG_\kappa}(b)}^{e_{\vG_\kappa}(a)}}\cdot \frac{\sum_{i=e_{\vG_\kappa}(b)}^{e_{\vG_\kappa}(a)}\lambda_{i,\kappa}}{e_{\vG_\kappa}(a)-e_{\vG_\kappa}(b)}L_{\kappa}(\vtheta)\notag\\&\le -2 \eta_{\kappa}(T) \cdot \min_{t\in[t_*,T]} \frac{\sum_{i=e_{\vG_\kappa}(b)}^{e_{\vG_\kappa}(a)}\lambda_{i,\kappa}}{e_{\vG_\kappa}(a)-e_{\vG_\kappa}(b)}\cdot L_{\kappa}(\vtheta).
\end{align}   

Therefore, we have that 
\begin{equation}
    L_{\kappa}(\vtheta)[t] \leq \exp\left(-2 \eta_{\kappa}(T) \cdot \min_{t\in[0,T]} \frac{\sum_{i=e_{\vG_\kappa}(b)}^{e_{\vG_\kappa}(a)}\lambda_{i,\kappa}}{e_{\vG_\kappa}(a)-e_{\vG_\kappa}(b)} (t-t_*)\right) \cdot L_{\kappa}(\vtheta)[t_*]
\end{equation} 
for all $t\in[t_*,T]$.

The proof for the other direction follows the same steps. 
\end{proof}

All the terms in Theorem \ref{speed} change with respect to the time $t$ during the training. Suppose for the two methods, $\eta_{\text{AD}}(T),\zeta_{\text{AD}}(T)$ and $\eta_{\text{FD}}(T),\zeta_{\text{FD}}(T)$ in Eq.~\ref{Constant} are close to each other respectively, we can use 
    $
    \frac{\sum_{i=e_{\vG_\kappa}(b)}^{e_{\vG_\kappa}(a)}\lambda_{i,\kappa}}{e_{\vG_\kappa}(a)-e_{\vG_\kappa}(b)}
    $ to quantify the training speed of AD and FD methods. In practical applications, the choice of $e_{\vG_\kappa}(b)$ may vary for different equations. The truncated entropy is related to this training speed as \[\frac{\log e_{\vG_{\kappa}(a)}}{e_{\vG_{\kappa}(a)}}H_{\vG_\kappa}(a)\approx\frac{\sum_{i=e_{\vG_{\kappa}}(b)}^{e_{\vG_{\kappa}}(a)}\lambda_{i,\kappa}}{e_{\vG_{\kappa}}(a)-e_{\vG_{\kappa}}(b)}.\]
This means that \textit{truncated entropy} can serve as an indicator of loss convergence speed.

Based on the relationship between the singular values of $\vA_{\text{AD}}$ and $\vA_{\text{FD}}$, as well as the eigenvalues of  $\vG_{\text{AD}}$ and $\vG_{\text{FD}}$, as demonstrated in Proposition~\ref{large} and Proposition~\ref{small}, we know that 
\begin{equation}
\begin{aligned}
H_{\vG_\text{AD}}(a) &\gg H_{\vG_\text{FD}}(a), \\
\frac{\log e_{\vG_\text{AD}(a)}}{e_{\vG_\text{AD}(a)}} &\ge \frac{\log e_{\vG_\text{FD}(a)}}{e_{\vG_\text{FD}(a)}}
\label{Prop3.1}
\end{aligned}
\end{equation}

Therefore, based on this, the speed of the AD method is faster than the FD due to Eq.~\ref{Prop3.1}. This can also be seen in the numerical results shown in Figure~\ref{PINNGD}(b) and Figure~\ref{PINNGD}(c).

\section{Numerical Results}\label{experiments}
In this section, we first apply both the random feature models (RFM) and two-layer neural networks (NN) methods discussed in Section~\ref{RFM} and Section~\ref{NN} to present numerical examples to compare AD and FD. Furthermore, we also conduct numerical experiments in random neural networks (deep neural network structure) and deep fully connected neural networks. All experiments are run on a single NVIDIA 3070Ti GPU.

\subsection{Two-layers Neural Networks}

In the case of the random feature model, the system precision is set to double, while for the neural network model, the system precision is set to float. To maintain consistency with our analysis, we train neural networks using the SGD optimizer with full batch training. We perform a learning rate sweep ranging from 1e-3 to 1e-4. All random feature models are structured as 2-layer fully connected NN, a $\sin(x)$ activation function, and with 
$\vw_{j} \sim \mathbb{U}\left([-1, 1]^{d}\right)$ and $b_{j} \sim \mathbb{U}\left([-1, 1]\right)$. All neural networks are 2-layer fully connected NN, a $\sin(x)$ activation function, and uniformly sample collocation points $x$ on the domain. Detailed experiment settings can be found in Appendix A.  In addition, we show numerical experiments concerning different grid sizes $h$, along with the corresponding training error curves and $L_2$ relative error curves in Appendix B.

\paragraph{\textbf{Poisson Equation in 2D}}

Consider the Poisson equation in 2D with the Dirichlet boundary condition over $\Omega=[0,1] \times[0,1]$, namely,
\begin{equation}
\begin{aligned}
-\Delta u(x, y)=f(x, y), \quad &(x, y) \in \Omega, \\
u(x,y) = 0 , \quad &(x, y) \in \partial \Omega.
\end{aligned}
\end{equation}
By choosing $f(x)=\pi^2\sin(\pi x)\sin(\pi y)$, we have the exact solution is $u(x,y) = \sin(\pi x)\sin(\pi y)$. The same finite difference scheme is used in RFM and NN in the FD case. In the neural network case, the loss function $L_{\text{PINN}}^{\text{AD}}(\vtheta)$ and $L_{\text{PINN}}^{\text{FD}}(\vtheta)$ are:

\begin{equation}
L_\text{PINN}^{\text{AD}}(\boldsymbol{\theta}) := L_{\text{F}}^{AD}(\theta) + \lambda L_\text{B}(\theta) =  \frac{1}{N}\sum_{i=1}^N |\Delta\phi(x_i;\boldsymbol{\theta}) - f(\vx)|^2 + \lambda \frac{1}{\hat{N}}\sum_{i=1}^{\widehat{N}} |\phi(\vy_i;\boldsymbol{\theta})|^2,
\end{equation}

\begin{equation}
\begin{aligned}
L_\text{PINN}^{\text{FD}}(\boldsymbol{\theta}) &:= L_{\text{F}}^{FD}(\theta) + \lambda L_\text{B}(\theta)  \\
&=  \frac{1}{N}\sum_{i=1}^N \left|\frac{\phi_{i+1,j}+\phi_{i-1,j} +\phi_{i,j-1}+\phi_{i,j+1}-\phi_{i,j}}{h^2}-f(x_i)\right|^2 + \lambda \frac{1}{\hat{N}}\sum_{i=1}^{\widehat{N}} |\phi(\vy_i;\boldsymbol{\theta})|^2,
\end{aligned}
\end{equation}
where $\lambda = 1$ and $h = \frac{1}{300}$ and $\phi_{i,j}:= \phi((x_i,y_j);\boldsymbol{\theta})$.

\begin{figure}[!ht]
    \centering
    \setcounter {subfigure} {0} (a){
    \includegraphics[scale=0.33]{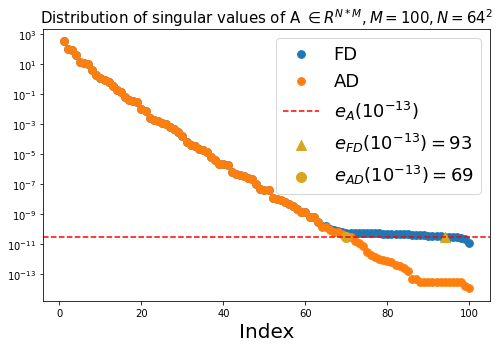}}
    \setcounter {subfigure} {0} (b){
    \includegraphics[scale=0.33]{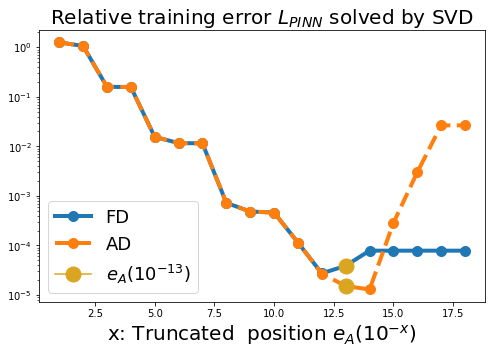}}\\
    \setcounter {subfigure} {0} (c){
    \includegraphics[scale=0.33]{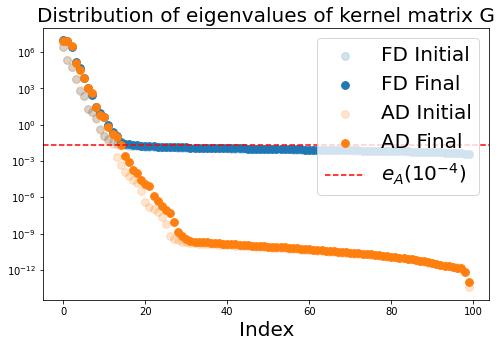}}
    \setcounter {subfigure} {0} (d){
        \includegraphics[scale=0.33]{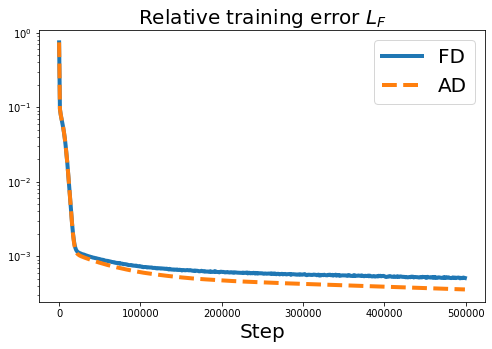}}
    \caption{\textbf{Poisson Equation in 2D.} \textbf{(a):} Distribution of singular values of the random feature matrix $\vA_k$ with the red dashed line indicating the effective cut-off number is $e_{\vG_k}(10^{-13})$. The \textit{truncated entropy} values are $H_{\vA_\text{FD}}(10^{-13}) = 0.3122$ and $H_{\vA_\text{AD}}(10^{-13}) = 0.3342$. \textbf{(b):} Relative training error of $L_\text{PINN}$, denoted as $\frac{\|\vA_k \va -\vf\|}{\|\vf\|}$, obtained using the truncated SVD method.  The horizontal axis represents the truncation position determined by the effective cut-off number. \textbf{(c):} Distribution of eigenvalues of kernel matrix $G$ with the red dashed line indicating the approximate convergence position $e_{\vG_k}(10^{-4})$ at the end of training. The \textit{truncated entropy} values are $H_{\vG_\text{FD}}(10^{-4}) = 0.3336$ and $H_{\vG_\text{AD}}(10^{-4}) = 0.4460$. \textbf{(d):}  Training curve of relative training error of the residual of PDE equation, i.e., $\frac{L_F}{\|f\|}$. (The training curve of the total loss $L_{\text{PINN}}$ is shown in Figure A.3.1.) }
    \label{Fig.2DPoisson}
\end{figure}

As shown in Figure~\ref{Fig.2DPoisson}, the distribution of singular values of matrix $\vA$ in the RFM and the distribution of eigenvalues of matrix $\vG$ in the NN exhibit the same relationship between the AD and FD methods as described in the previous analysis: $\vA_{\text{AD}}$ and $\vA_{\text{FD}}$ share the similar large singular values, and for small singulars $\vA_{\text{FD}}$ are larger than $\vA_{\text{AD}}$.  

By comparing with the 1D Poisson equation (see Figure~\ref{PINNGD}(b)), the difference in training error between AD and FD is small. This can be explained by the small difference in truncated entropy between AD and FD as shown in Figure~\ref{Fig.2DPoisson}(c) where $H_{\vG_\text{FD}}(10^{-4}) = 0.3336$ and $H_{\vG_\text{AD}}(10^{-4}) = 0.4460$. In this example, AD still outperforms FD in training.

\paragraph{\textbf{Biharmonic Equation}}

We consider the fourth-order two-point boundary value problem 
\begin{equation}
\displaystyle   \frac{\D^4u(x)}{\D^4x} = \exp(x)
\end{equation} 
with boundary conditions $u(-1) = u(1) = 0$ and $u^{\prime}(-1) = u^{\prime}(1) = 0$. The exact solution is $u(x) = c_0 + c_1x+c_2x^2+c_3x^3+\exp(x)$ with $c_0=-\frac{5}{12}e^{-1}-\frac{1}{4}e, c_1=\frac{1}{2}e^{-1}-\frac{1}{2}e, c_2=\frac{1}{4}e^{-1}-\frac{1}{4}e, c_3=-\frac{1}{3}e^{-1}$. In 2D Poisson equation, the loss functions $L_{\text{PINN}}^{\text{AD}}(\vtheta)$ and $L_{\text{PINN}}^{\text{FD}}(\vtheta)$ are:

\begin{equation}
\begin{aligned}
L_\text{PINN}^{\text{AD}}(\boldsymbol{\theta}) &:= L_{\text{F}}^{AD}(\theta) + \lambda L_\text{B}(\theta) \\ &=  \frac{1}{N}\sum_{i=1}^N |\phi_{xxxx}(x_i;\boldsymbol{\theta}) - f(\vx)|^2 + \lambda \frac{1}{\hat{N}}(\sum_{i=1}^{\widehat{N}} |\phi(y_i;\boldsymbol{\theta})|^2 +\sum_{i=1}^{\widehat{N}} |\phi_x(y_i;\boldsymbol{\theta})|^2  ),
\end{aligned}
\end{equation}

\begin{equation}
\begin{aligned}
L_\text{PINN}^{\text{FD}}(\boldsymbol{\theta}) &:= L_{\text{F}}^{FD}(\theta) + \lambda L_\text{B}(\theta)  \\
=&  \frac{1}{N}\sum_{i=1}^N \left|\frac{\phi_{i-2}-4\phi_{i-1} +6\phi_{i}-4\phi_{i+1}+\phi_{i+2}}{h^4}-f(x_i)\right|^2 \\ &+ \lambda\frac{1}{\hat{N}}(\sum_{i=1}^{\widehat{N}} |\phi(y_i;\boldsymbol{\theta})|^2+\sum_{i=1}^{\widehat{N}} |\phi_x(y_i;\boldsymbol{\theta})|^2  ),
\end{aligned}
\end{equation}
where $\lambda = 100$ and $h = \frac{2}{64}$ and $\phi_{i}:= \phi(x_i;\boldsymbol{\theta})$.

\begin{figure}[!ht]
    \centering
    \setcounter {subfigure} {0} (a){
    \includegraphics[scale=0.33]{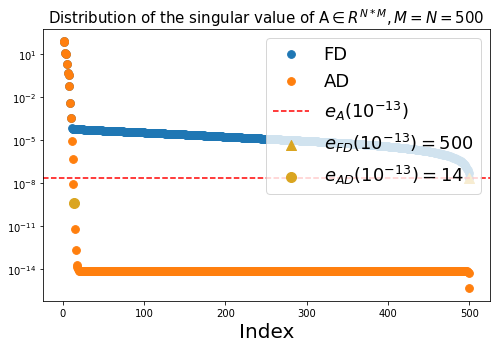}}
    \setcounter {subfigure} {0} (b){
    \includegraphics[scale=0.33]{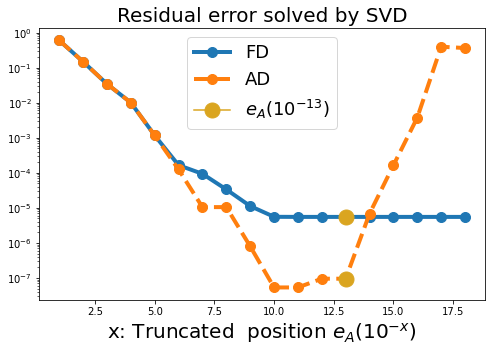}}
    \setcounter {subfigure} {0} (c){
    \includegraphics[scale=0.33]{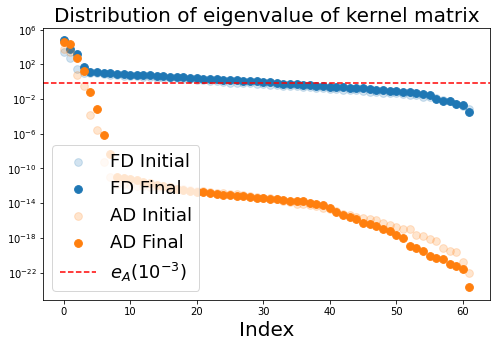}}
    \setcounter {subfigure} {0} (d){
        \includegraphics[scale=0.33]{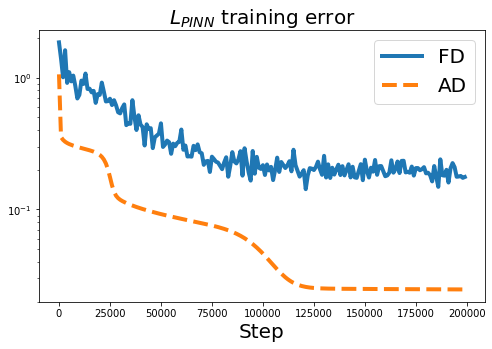}}
    \caption{\textbf{Biharmonic Equation.} \textbf{(a):} Distribution of singular values of the random feature $\vA_k$ with the red dashed line indicating the effective cut-off number is $e_{\vG_k}(10^{-13})$. The \textit{truncated entropy} values are $H_{\vA_\text{FD}}(10^{-13}) = 0.1895$ and $H_{\vA_\text{AD}}(10^{-13}) = 0.4460$. \textbf{(b):} Relative training error of $L_\text{PINN}$, denoted as $\frac{\|\vA_k \va -\vf\|}{\|\vf\|}$, obtained using the truncated SVD method.  The horizontal axis represents the truncation position determined by the effective cut-off number. \textbf{(c):} Distribution of eigenvalues of kernel matrix $G$ with the red dashed line indicating the approximate convergence position $e_{\vG_k}(10^{-2})$ at the end of training. The \textit{truncated entropy} values are $H_{\vG_\text{FD}}(10^{-2}) = 0.5058$ and $H_{\vG_\text{AD}}(10^{-2}) = 0.6720$. \textbf{(d):} Training curve of relative training error of the residual of PDE equation, i.e., $\frac{L_F}{\|f\|}$. (The training curve of the total loss $L_{\text{PINN}}$ is shown in Figure A.4.1.) }
    \label{Fig.Biharmonic}
\end{figure}

Due to the high order derivatives, the difference in the singular values distribution of $\vA$ and eigenvalue distribution of $\vG$ between AD and FD becomes more pronounced, as shown in Figure~\ref{Fig.Biharmonic}(a)(c). This leads to a larger difference in truncated entropy, i.e., $H_{\vA_\text{FD}}(10^{-13}) = 0.1895 < H_{\vA_\text{AD}}(10^{-13}) = 0.4460$ and  $H_{\vG_\text{FD}}(10^{-2}) = 0.5058<H_{\vG_\text{AD}}(10^{-2}) = 0.6720$. Thus in this example, the training error of AD converges faster than FD, as shown in Figure~\ref{Fig.Biharmonic}(b)(d).


\subsection{Deep Neural Networks}
 We further perform numerical experiments to validate that the different training behaviors of AD and FD persist in deep neural network architectures. We use 1D Poisson equation as shown in Eq.~\ref{Eqn.1DPoisson} as a numerical example to compare AD and FD on deep neural network structure.

\textbf{Random Neural Network.} Here we chose $N = 500$. The random neural network is structured as 4-layer fully connected NN with 50 hidden neurons ([1,50,50,50,1]), a $\tanh(x)$ activation function, and with 
$\vw_{j} \sim \mathbb{U}\left([-0.1, 0.1]^{d}\right)$ and $b_{j} \sim \mathbb{U}\left([-0.1, 0.1]\right)$.
Here, we also take into account the higher-order Five-Point Stencil method in Eq.~\ref{FD_Five}.

\begin{equation}
\Delta_\text{FD (five-point)} = \frac{-\phi_{\theta}(x-2h)+16\phi_{\theta}(x-h)-30\phi_{\theta}(x)+16\phi_{\theta}(x+h) - \phi_{\theta}(x+2h)}{12h^2}.
\label{FD_Five}
\end{equation}

\begin{figure}[!ht]
    \centering
    \setcounter {subfigure} {0} (a){
    \includegraphics[scale=0.33]{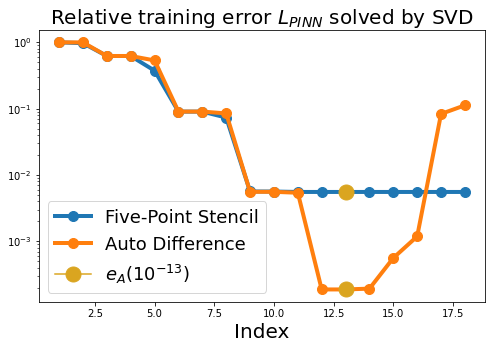}}
    \setcounter {subfigure} {0} (b){
    \includegraphics[scale=0.33]{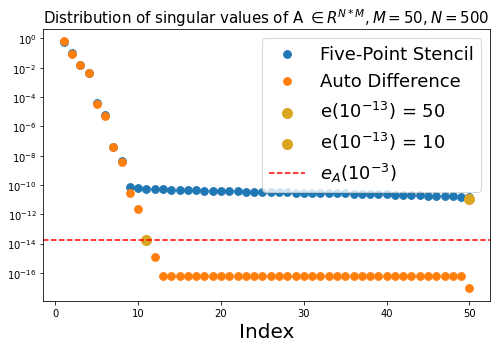}}
    \caption{\textbf{Random neural network.} \textbf{(a):} Distribution of singular values of the random feature $\vA_k$ with the red dashed line indicating the effective cut-off number is $e_{\vG_k}(10^{-13})$. The \textit{truncated entropy} values are $H_{\vA_\text{FD}}(10^{-13}) = 0.1293$ and $H_{\vA_\text{AD}}(10^{-13}) = 0.2197$. \textbf{(b):} Relative training error of $L_\text{PINN}$, denoted as $\frac{\|\vA_k \va -\vf\|}{\|\vf\|}$, obtained using the truncated SVD method.  The horizontal axis represents the truncation position determined by the effective cut-off number.}
    \label{Fig.RNN}
\end{figure}

The results are shown in Figure~\ref{Fig.RNN}. It shows that in random neural networks, the relationship between the singular values of $\vA_{\text{AD}}$ and  $\vA_{\text{FD}}$ remains consistent with the results of the random feature models (see Figure~\ref{Fig.RNN}(a)). The training error of AD is smaller than that of FD ( see Figure~\ref{Fig.RNN}(b)).

\textbf{Deep neural network.} The deep neural network is structured as 4-layer fully connected NN with 50 hidden neurons ([1,50,50,50,1]), a $\tanh(x)$ activation function. We select the number of grid points $N = 100$ with the Adam optimization method using full batch training. And we also discuss the impact of different finite difference methods on the results.

\begin{figure}[!ht]
    \centering
    \setcounter {subfigure} 0(a){
    \includegraphics[scale=0.35]{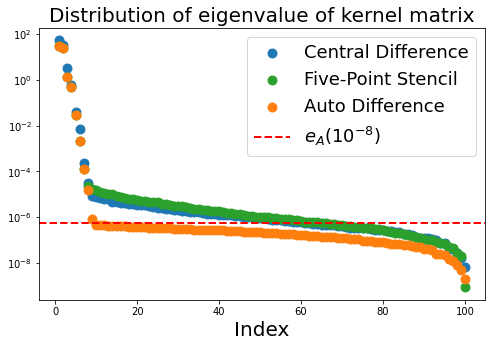}} \\
    \setcounter {subfigure} 0(b){
    \includegraphics[scale=0.35]{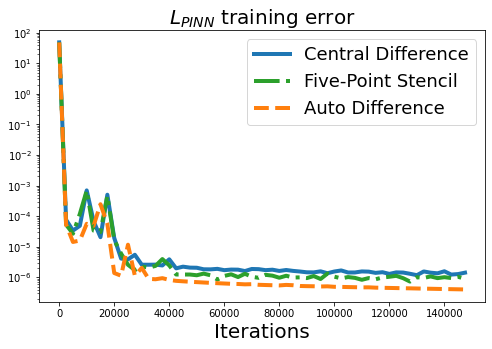}}
    \setcounter {subfigure} 0(c){
    \includegraphics[scale=0.35]{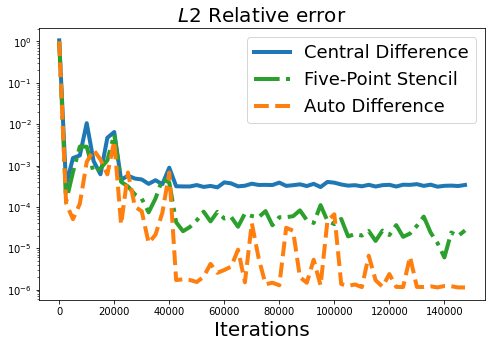}}
    \caption{\textbf{(a):} Distribution of eigenvalues of kernel matrix $G$ with the red dashed line indicating the approximate convergence position $e_{\vG_k}(10^{-8})$ at the end of training. \textit{Truncated entropy} values are $H_{\vG_\text{FD (central difference)}}(10^{-8}) = 0.2043$, $H_{\vG_\text{FD (five-point)}}(10^{-8}) = 0.1911$  and $H_{\vG_\text{AD}}(10^{-8}) = 0.2410$. \textbf{(b):} Training curve of relative training error $\frac{L_F}{\|f\|}$. \textbf{(c):} Training curve of the Relative L2 error.}
    \label{Fig.DNN}
\end{figure}

The results is shown in Figure~\ref{Fig.DNN}. For the numerical results of deep neural networks, after a certain stage of training, AD converges faster than FD. Furthermore, the final $L_2$ relative error is better for AD compared to FD as shown in Figure~\ref{Fig.DNN}(c). For different finite difference methods, the primary impact is on the final Relative $L_2$ error. As shown in Figure~\ref{Fig.DNN}(c), the Relative $L_2$ error of the higher-order Five-Point Stencil method is better than that of the central difference scheme, which is mainly caused by the generalization error. However, our article primarily focuses on the comparison of training errors, we will not elaborate further on this aspect. This numerical example demonstrates that our findings can be extended to deep neural network structures.

\subsection{Nonlinear case}
\begin{figure}[!ht]
    \centering
    \setcounter {subfigure} 0(a){
    \includegraphics[scale=0.35]{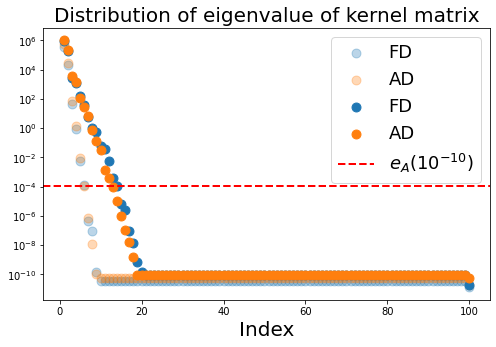}}
    \setcounter {subfigure} 0(b){
    \includegraphics[scale=0.35]{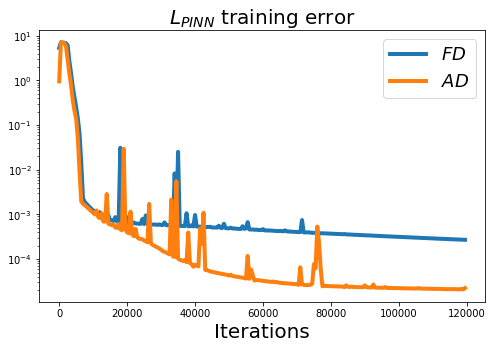}}
    \caption{\textbf{Allen-Cahn Equation.} \textbf{(a)}: Distribution of eigenvalues of kernel matrix $G$ with the red dashed line indicating the approximate convergence position $e_{\vG_k}(10^{-10})$ at the end of training. The \textit{truncated entropy} values at final are $H_{\vG_\text{FD}}(10^{-10}) = 0.1810$ and $H_{\vG_\text{AD}}(10^{-10}) = 0.2059$. \textbf{(b):} Training curve of relative training error of the residual of PDE equation.  }
    \label{Fig.AC}
\end{figure}

Here, we also use an experiment involving a nonlinear equation to demonstrate that our conclusions are equally applicable to nonlinear equations. Specifically, we consider the steady-state solution of the one-dimensional Allen-Cahn Equation as follows.
\begin{equation}
\left\{\begin{array}{l}
\epsilon u^{\prime \prime}(x) + \frac{u^3 - u}{\epsilon}=0, \quad x \in[0,1] \\
u(0)=-1, \quad u(1)=1,
\end{array}\right.
\end{equation}
where we set $\epsilon=0.1$. The equation has the steady state solution:
\begin{equation}
    u(x) = \tanh\left(\frac{x-0.5}{\sqrt{2}\epsilon}\right).
\end{equation}
Here, we handle the Laplacian operator in the equation using the central difference scheme and automatic differentiation, respectively. We use deep neural network to conduct the experiments, the results are as shown in Figure~\ref{Fig.AC}. 
From Figure~\ref{Fig.AC}(b), it can be observed that the training convergence of the AD method is still faster than that of the FD method, which is consistent with our conclusions. 

\begin{rmk}
Compared to AD, FD is often easier to implement and computationally less demanding. 
Our work focuses on comparing AD and FD  the perspective of training error, which is different from the perspective of training time.

Based on our experiments, when working with lower-order differential operators, such as the Laplacian, the training time for AD and FD is almost identical. For higher-order differential operators, such as the fourth-order biharmonic equation discussed in the paper, the training time for FD is noticeably faster than for AD.

Since our primary focus is on training error, it is important to clarify that our comparison of training convergence speed is based on the rate at which the training loss decreases over the same number of epochs. We observed that during the later stages of training, FD methods encounter difficulties in further reducing the training loss due to the presence of numerous negligible small eigenvalues in the FD kernel. As a result, even when the training duration is the same, the AD method consistently achieves a lower training loss than FD.
\end{rmk}


\section{Conclusions and Discussions}
In this paper, we study the performance of AD and FD methods in training neural network to solve PDEs. Neural network structures include random feature models and two-layer neural networks. We believe that this is the first paper to compare AD and FD from a training perspective theoretically and numerically. We analyze the singular values of $\vA_{\text{AD}}$ and $\vA_{\text{FD}}$ in random feature models. Our analysis of the singular values shows that $\vA_{\text{AD}}$ and $\vA_{\text{FD}}$ share the similar large singular value (see Proposition~\ref{large}), and for the small singular values, $\vA_{\text{AD}}$ are smaller than $\vA_{\text{FD}}$ (see Proposition~\ref{small}). Through Theorem~\ref{speed}, we describe the influence of eigenvalues of kernel $\vG$ on training error. In addition, we introduce the concepts of effective cut-off number and truncated entropy, observing that larger truncated entropy corresponds to smaller training error. The truncated entropy of AD is larger than that of FD, predicting that AD outperforms FD methods in the training process. This analysis can be extended to general neural network training scenarios, with truncated entropy as an indicator of loss convergence speed.

Our analysis is based on PINNs with random feature networks and two-layer neural networks. We chose these models as a starting point for our investigation due to their simplicity and ease of analysis. The same analytical techniques and phenomena can be extended to deeper neural network structures, such as random neural networks and deeper fully connected neural networks. Exploring the properties of solving PDEs using different neural network architectures such as convolutional neural networks (CNN) and transformers can be considered as future work. In neural network methods for solving PDEs, the error can be divided into three components: approximation error, generalization error, and training error. Our paper focuses on the training error introduced by differential methods, distinct from approximation and generalization errors. Note that the relative error against a ground truth solution decreases as the training error decreases. This provides insights into how to train the loss function when derivative information is included effectively. Since incorporating preconditioners into optimization algorithms could accelerate the convergence of training errors, further research can be done to investigate the most suitable preconditioners for different PDE types and neural network architectures.

\section*{Acknowledgements}
\textbf{Funding} Y.Y. and W.H. was supported by National Institute of General Medical Sciences through grant 1R35GM146894. The work of Y.X. was supported by the Project of Hetao Shenzhen-HKUST Innovation Cooperation Zone HZQB-KCZYB-2020083.

\section*{Data Availability}
Data availability The data that support the findings of this study are available from the corresponding author upon reasonable request.

\section*{Declarations}
\textbf{Conflict of interest} The authors have no relevant financial or non-financial interests to disclose.


 \bibliographystyle{elsarticle-num} 
 \bibliography{references}

\appendix
\section*{Appendix}

\section{Experimental Details}
In this section, we include more details about the numerical experiments shown in the main text.

\subsection{Training setting}
In the case of the random feature model, the system precision is set to double, while for the neural network model, the system precision is set to float. We train neural networks using the SGD optimizer with full batch training. We perform a learning rate sweep ranging from 1e-3 to 1e-4. All random feature models are structured as 2-layer fully connected NN,a $\sin(x)$ activation function, and with 
$\vw_{j} \sim \mathbb{U}\left([-1, 1]^{d}\right)$ and $b_{j} \sim \mathbb{U}\left([-1, 1]\right)$. All neural networks are 2-layer fully connected NN, a $\sin(x)$ activation function, and uniformly sample collocation points $x$ on the domain. All experiments are run on a single NVIDIA 3070Ti GPU.

\subsection{Poisson Equation in 1D}

\textbf{Equation.}
\begin{equation}
\left\{\begin{array}{l}
u_{xx}(x)=f(x), \quad x \in [-1,1], \\
u(-1) = u(1) = 0.
\end{array}\right.
\tag{A.2.1}
\end{equation}
By choosing $f(x)=-\pi^2\sin(\pi x)$, we have the exact solution is $u(x) = \sin(\pi x)$.

\textbf{Random feature model.} We include extra distribution of singular values of $A_{\text{AD}}$ and $A_{\text{FD}}$ for different $M=N$ as shown in Figure~\ref{Residual svd1} in the main context. The activation function is $\sin(x)$. As shown in Figure~\ref{Fig.Appendix2.1}, for for varying numbers of hidden neurons $M$ and grid points $N$, the relationship between the singular values of $\vA_{\text{AD}}$ and $\vA_{\text{FD}}$ persists as described in Proposition~\ref{large} and Proposition~\ref{small}.
\renewcommand{\thefigure}{A.2.1}
\begin{figure}[!ht]
    \centering
    \includegraphics[scale=0.48]{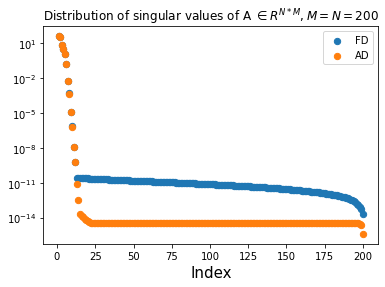}
    \includegraphics[scale=0.48]{RFM_Singular_eigen_M300.png}
    \includegraphics[scale=0.48]{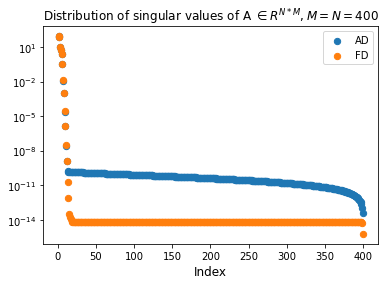}
    \includegraphics[scale=0.48]{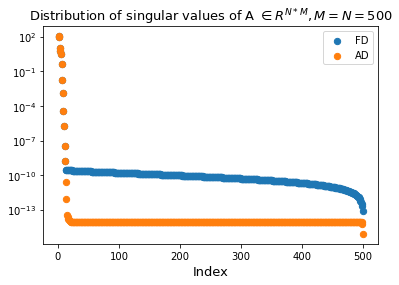}
    \caption{Distribution of singular values of $\vA_{\text{AD}}$ and $\vA_{\text{FD}}$ for different values of $M=N$.}
\label{Fig.Appendix2.1}
\end{figure}

\textbf{Neural Network.} As shown in Figure~\ref{PINNGD} in the main context. Here, we use 2-layer fully connected NN with $100$ neurons per layer, a $\sin(x)$ activation function, and uniformly $100$ sample collocation points. The loss function for AD and FD are: 

\begin{equation}
L_\text{PINN}^{\text{AD}}(\boldsymbol{\theta}) := L_{\text{F}}^{AD}(\theta) + \lambda L_\text{B}(\theta) =  \frac{1}{N}\sum_{i=1}^N |\Delta\phi(\vx_i;\boldsymbol{\theta}) - f(\vx)|^2 + \lambda \frac{1}{\hat{N}}\sum_{i=1}^{\widehat{N}} |\phi(\vy_i;\boldsymbol{\theta})|^2,
\tag{A.2.2}
\end{equation}

\begin{equation}
\begin{aligned}
L_\text{PINN}^{\text{FD}}(\boldsymbol{\theta}) &:= L_{\text{F}}^{FD}(\theta) + \lambda L_\text{B}(\theta)  \\
&=  \frac{1}{N}\sum_{i=1}^N \left|\frac{\phi(x_{i-1};\boldsymbol{\theta})+\phi(x_{i+1};\boldsymbol{\theta})-2\phi(x_i;\boldsymbol{\theta})}{h^2}-f(x_i)\right|^2 + \lambda \frac{1}{\hat{N}}\sum_{i=1}^{\widehat{N}} |\phi(\vy_i;\boldsymbol{\theta})|^2,
\end{aligned}
\tag{A.2.3}
\end{equation}
where $\lambda = 1$ and $h = \frac{2}{100}$.

\subsection{Poisson Equation in 2D.}

\textbf{Equation.}
\begin{equation}
\left\{\begin{array}{l}
\Delta u(x, y)=f(x, y), \quad(x, y) \in \Omega, \\
u(x,y) = 0 , \quad(x, y) \in \partial \Omega.
\end{array}\right.
\tag{A.3.1}
\end{equation}
By choosing $f(x)=-\pi^2\sin(\pi x)\sin(\pi y)$, we have the exact solution is $u(x,y) = \sin(\pi x)\sin(\pi y) \in [0,1]\times [0,1]$.

\textbf{Random Feature Model.}  In this case the $M = 100, N = 64\times 64$ $\vA_\text{FD} = \vC \cdot \vA_0$.
$$
C=\frac{1}{h^2}\left[\begin{array}{ccccccc}
D & -I & 0 & 0 & 0 & \cdots & 0 \\
-I & D & -I & 0 & 0 & \cdots & 0 \\
0 & -I & D & -I & 0 & \cdots & 0 \\
\vdots & \ddots & \ddots & \ddots & \ddots & \ddots & \vdots \\
0 & \cdots & 0 & -I & D & -I & 0 \\
0 & \cdots & \cdots & 0 & -I & D & -I \\
0 & \cdots & \cdots & \cdots & 0 & -I & D
\end{array}\right]
$$
$I$ is the $64 \times 64$ identity matrix, and $D$, also $64 \times 64$, is given by:
$$
D=\left[\begin{array}{ccccccc}
4 & -1 & 0 & 0 & 0 & \cdots & 0 \\
-1 & 4 & -1 & 0 & 0 & \cdots & 0 \\
0 & -1 & 4 & -1 & 0 & \cdots & 0 \\
\vdots & \ddots & \ddots & \ddots & \ddots & \ddots & \vdots \\
0 & \cdots & 0 & -1 & 4 & -1 & 0 \\
0 & \cdots & \cdots & 0 & -1 & 4 & -1 \\
0 & \cdots & \cdots & \cdots & 0 & -1 & 4
\end{array}\right].
$$

\textbf{Neural Network.} Figure~\ref{Fig.2DPoisson} in the main context shows the convergence curve of the residual on the PDE equation $L_{\text{F}}$. Figure A.3.1 illustrates the convergence of the overall loss $L_{\text{PINN}}$. The convergence curves of these two are similar.  Here, we use 2-layer fully connected NN with $100$ neurons per layer, a $\sin(x)$ activation function, and uniformly $N 
 = 64\times64$ sample collocation points. 

\renewcommand{\thefigure}{A.3.1}
\begin{figure}[!ht]
    \centering
    \includegraphics[scale=0.55]{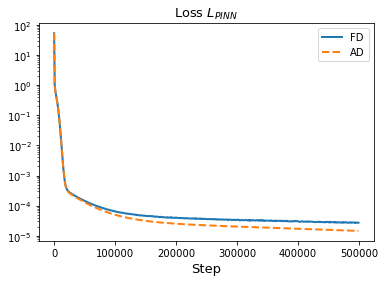}
    \caption{Training curve of the total loss function $L_{\text{PINN}}$.}
\end{figure}

\subsection{Biharmonic Equation.}

\textbf{Equation}
\begin{equation}
    \frac{\D^4u(x)}{\D^4x} = \exp(x),
\tag{A.4.1}
\end{equation}
with boundary conditions $u(-1) = u(1) = 0$ and $u^{\prime}(-1) = u^{\prime}(1) = 0$. The exact solution is $u(x) = c_0 + c_1x+c_2x^2+c_3x^3+\exp(x)$ with $c_1=-\frac{5}{12}e^{-1}-\frac{1}{4}e, c_2=\frac{1}{2}e^{-1}-\frac{1}{2}e, c_3=\frac{1}{4}e^{-1}-\frac{1}{4}e, c_3=-\frac{1}{3}e^{-1}$.

\textbf{Random feature model.} In this case the $M =N =500$ $\vA_\text{FD} = \vC \cdot \vA_0$

$$
C=\frac{1}{h^4}\left[\begin{array}{ccccccc}
-4 & 6 & -4 & 0 & 0  & \cdots & 0 \\
1 & -4 & 6 & -4 & 1  & \cdots & 0 \\
0 & 1 & -4 & 6 & -4 &  \cdots & 0 \\
\vdots & \ddots & \ddots & \ddots & \ddots & \ddots & \vdots \\
0 & \cdots & 4 & -6 & 4 & 1 & 0 \\
0 & \cdots & \cdots & 4 & -6 & 4 & 1 \\
0 & \cdots & \cdots & \cdots & 4 &-6 & 4
\end{array}\right]
$$

\textbf{Neural Network.} Figure~\ref{Fig.Biharmonic} in the main context shows the convergence curve of the residual on the PDE equation $L_{\text{F}}$. Figure A.4.1 illustrates the convergence of the overall loss $L_{\text{PINN}}$. The convergence curves of these two are similar. Here, we use 2-layer fully connected NN with $100$ neurons per layer, a $\sin(x)$ activation function, and uniformly $N 
 = 64\times64$ sample collocation points. 

\renewcommand{\thefigure}{A.4.1}
\begin{figure}[ht!]
    \centering
    \includegraphics[scale=0.55]{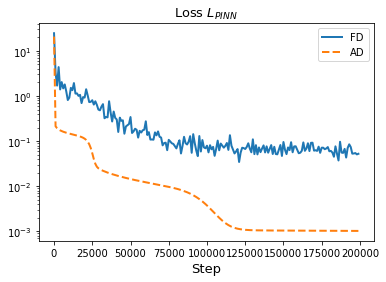}
    \caption{Training curve of the total loss function $L_{\text{PINN}}$.}
\end{figure}

\section{Additional Numerical Experiments.}
In this section, we present additional experimental results showcasing the outcomes in the case of the 1D Poisson Equation. For different grid sizes $h$, the results include the loss curve and the corresponding Relative L2 error curve.

\subsection{Different Grid Size h.}

\renewcommand{\thefigure}{B.1}
\begin{figure}[!ht]
    \centering
    \setcounter {subfigure} 0(a){
    \includegraphics[scale=0.43]{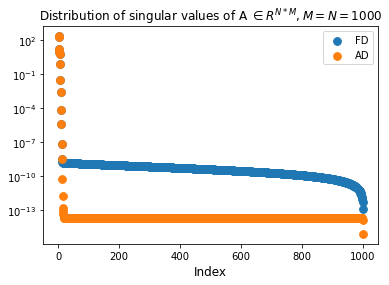}}
    \setcounter {subfigure} 0(b){
    \includegraphics[scale=0.43]
    {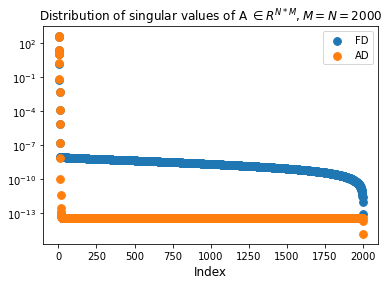}}
    \caption{Distribution of singular values for $\boldsymbol{A}_{\text{FD}}$ and $\boldsymbol{A}_{\text{AD}}$ for solving 1D Poisson equation using Random Feature Method with activation function $\sigma(x)=\sin(x)$ with vary  
Number of Sample Points $N$ and Number of Neurons $M$. \textbf{(a):} $M=N=1000$. \textbf{(b):} $M=N=2000$.}
    \label{Fig.differenth}
\end{figure}

As shown in Figure~\ref{Fig.differenth}, we extend the analysis of eigenvalue distribution for the 1D Poisson equation with $M=N$ under the Random Feature Model. Results corresponding to Proposition 1 and Proposition 2 in Section 3.1 are applicable across various scenarios.

\subsection{Loss Curve and Relative L2 Error}

Figure~\ref{Fig.lcurverelativecurve} presents the convergence behavior of the loss curve and the corresponding relative L2 error for the 1D Poisson equation with $M=N$. It shows that AD consistently achieves lower training errors and relative errors compared to FD, demonstrating its superior performance in this setting.

\renewcommand{\thefigure}{B.2}
\begin{figure}[!ht]
    \centering
    \setcounter {subfigure} 0(a.1){
    \includegraphics[scale=0.41]{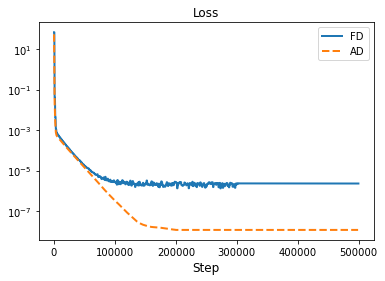}}
    \setcounter {subfigure} 0(a.2){
    \includegraphics[scale=0.41]{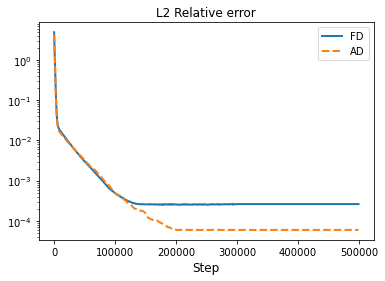}}
    \setcounter {subfigure} 0(b.1){
    \includegraphics[scale=0.41]{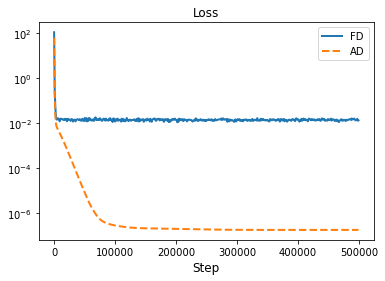}}
    \setcounter {subfigure} 0(b.2){
    \includegraphics[scale=0.41]{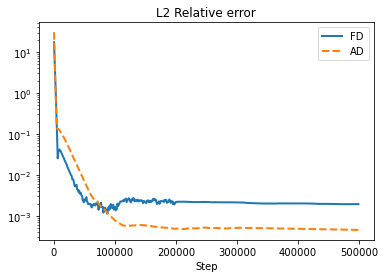}}
    \setcounter {subfigure} 0(c.1){
    \includegraphics[scale=0.41]{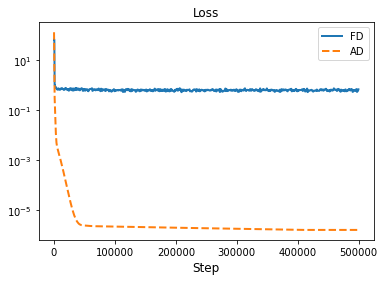}}
    \setcounter {subfigure} 0(c.2){
    \includegraphics[scale=0.41]{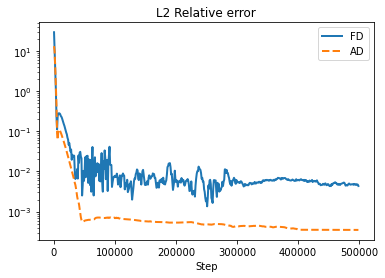}}
    \caption{Training curve of training error $L$ and its corresponding relative $L_2$ error. \textbf{(a)}: $M=N=100$. \textbf{(b)}:$M=N=500$. \textbf{(c)}:$M=N=1000$}
    \label{Fig.lcurverelativecurve}
\end{figure}




\end{document}